\documentclass[sn-mathphys]{sn-jnl}

\usepackage{amssymb}
\usepackage{amsmath} 
\usepackage{placeins}
\usepackage[utf8]{inputenc}
\usepackage[T1]{fontenc}
\usepackage{url}
\usepackage{multirow}
\usepackage[skip=0.5\baselineskip]{caption}
\usepackage{hyperref}
\usepackage{subcaption}
\usepackage{mdframed}
\usepackage{pifont}

\usepackage{lineno}

\newcommand{\sinergym}{\textsc{Sinergym}}
\newcommand{\boptest}{\textsc{Boptest}}
\newcommand{\energym}{\textsc{Energym}}
\newcommand{\rltestbed}{\textsc{RL-testbed}}

\newcommand{\cmark}{\ding{51}}%
\newcommand{\xmark}{\ding{55}}%

\newcommand{\fivezone}{\textsc{5ZoneAutoDXVAV}}
\newcommand{\datacenter}{\textsc{2ZoneDataCenterHVAC}}

\newcommand{\edited}[1]{#1} 

\hypersetup{
    colorlinks=true
}

\graphicspath{{img/}}

\jyear{2022}%

\begin{document}


\title[An experimental evaluation of DRL algorithms for HVAC control]{An experimental evaluation of Deep Reinforcement Learning algorithms for HVAC control}


\author*[1]{\fnm{Antonio} \sur{Manjavacas}}\email{manjavacas@ugr.es}
\author[1]{\fnm{Alejandro} \sur{Campoy-Nieves}}\email{alejandrocn7@ugr.es}
\author[1]{\fnm{Javier} \sur{Jiménez-Raboso}}\email{jajimer@correo.ugr.es}
\author[1]{\fnm{Miguel} \sur{Molina-Solana}}\email{miguelmolina@ugr.es}
\author[1]{\fnm{Juan} \sur{Gómez-Romero}}\email{jgomez@decsai.ugr.es}

\affil[1]{\orgdiv{Department of Computer Science and Artificial Intelligence}, \orgname{Universidad de Granada}, \orgaddress{\city{Granada}, \postcode{18071}, \country{Spain}}}


\abstract{Heating, Ventilation, and Air Conditioning (HVAC) systems are a major driver of energy consumption in commercial and residential buildings. Recent studies have shown that Deep Reinforcement Learning (DRL) algorithms can outperform traditional reactive controllers. However, DRL-based solutions are generally designed for ad hoc setups and lack standardization for comparison. To fill this gap, this paper provides a critical and reproducible evaluation, in terms of comfort and energy consumption, of several state-of-the-art DRL algorithms for HVAC control. The study examines the controllers' robustness, adaptability, and trade-off between optimization goals by using the \sinergym{} framework. The results obtained confirm the potential of DRL algorithms, such as SAC and TD3, in complex scenarios and reveal several challenges related to generalization and incremental learning.}




\maketitle

\section{Introduction}
\label{sec:1}

In recent decades, both global warming and climate change have been significantly spurred by the growth in energy demand from residential and non-residential buildings. The Global Alliance for Buildings and Construction reported that, in 2021, the building and construction sector was responsible for more than one third of the global energy consumption \cite{iea2021,globalabc2021} ---despite the reduction in commercial and industrial activities caused by the COVID-19 pandemic---. According to the International Energy Agency \cite{iea2021}, buildings are responsible for 17\% and 10\% of global direct and indirect CO$_2$ emissions, respectively.  

Heating, Ventilation, and Air Conditioning (HVAC) systems are one of the main sources of energy consumption in buildings, representing more than 50\% of their associated energy demand in developed countries \cite{perez2008}. This consumption is potentially affected by a lack of precise control over these systems, whose proper and efficient operation is vital to ensure energy savings \cite{wang2020a,mawson2021,gholamzadehmir2020}.

Many strategies have been implemented to improve HVAC control efficiency. In fact, most common control solutions are based on Rule-Based Controllers (RBC) combined with Proportional-Integrative-Derivative (PID) control \cite{geng1993}, which stand out for their simplicity and efficiency \cite{borase2021review}. By using heuristic rules, these controllers can guarantee proper comfort temperature ranges while reducing energy consumption \cite{ashrae2021}. However, this type of controllers are far from offering optimal behavior, as they are mostly reactive, hardly cover the complexity of environments influenced by many variables ---and therefore, many rules---, and are guided by fixed and predetermined control sequences that heavily rely on expert knowledge \cite{wang2020b,salsbury2005}. Generally, RBCs lack scalability, as they do not address building energy optimization on a global level, but on a local one. This is because a control where so many variables are considered would result in an overly complex RBC, being practically impossible to generalize their rules at a building level \cite{privara2013building,serale2018model}.

Model-based solutions, such as Model Predictive Control (MPC) \cite{yao2021,salakij2016,morari1999}, are an alternative to reactive controllers. These controllers use physical models of buildings to simulate their thermal dynamics and analytically derive optimal HVAC control. Usually, MPC consider not only the characteristics of the environment but also other constraints and contextual data such as occupancy or weather. This ability to characterize and predict environmental conditions makes MPC outperform reactive controllers \cite{yao2021,kumpel2021,efheij2019}. Nevertheless, MPC also suffers from certain limitations. In addition to the high computational power it demands, precise system calibration is necessary to achieve the expected performance, which is not easily scalable since each building is unique in its layout and thermodynamic properties \cite{gomez2019}. This poses a challenge in terms of cost-effectiveness, as many factors must be considered: materials, occupancy, end-use, location, orientation, etc. \edited{Consequently, this has resulted in a relatively small number of buildings currently implementing MPC strategies compared to `if-then-else', `on/off' or `bang-bang' RBC controllers, as well as PID where digital control and variable frequency drives are available} \cite{serale2018}.

Given the shortcomings of these methods, Reinforcement Learning (RL) has been recently proposed as a viable alternative for complex control problems. RL is a computational learning method focused on the interaction of an agent with its environment, either real or simulated. This is an iterative learning method, based on trial and error, where a reward function makes the agent lean towards preferable actions or states. Therefore, the agent's goal will be to discover which actions lead to the maximization of the expected reward \cite{sutton2018}. The combination of RL with deep neural networks has led to a growing application of Deep Reinforcement Learning (DRL) in numerous domains \cite{mnih2015,gibney2016,gupta2021}, including HVAC control \cite{barrett2015,azuatalam2020,perera2021,yang2021,fu2022}. Accordingly, DRL can learn sophisticated control strategies from data \cite{biemann2021}, generally obtained from building simulations, while using more computationally efficient building models than MPCs \cite{deng2022}. 

Nevertheless, as highlighted in \cite{vazquez2019a, findeis2022}, most of the DRL proposals for HVAC control in the literature pick one or few algorithms without substantial motivation, lack a comprehensive analysis of them under controlled and assorted conditions, and cannot be easily reproduced. Motivated by this gap, this paper performs a comprehensive experimental evaluation of state-of-the-art DRL algorithms applied to HVAC control. The main contribution of this work is offering insights on which algorithms are most promising in different building energy control scenarios, and what possibilities arise for their further improvement. To this aim, we focus on algorithms performance (evaluation metrics, comfort-consumption trade-off, convergence, etc.), but also in their robustness against changing conditions and capabilities for transference to different scenarios. The study relies on \sinergym{}\footnote{\url{https://github.com/ugr-sail/sinergym}}, an open-source building simulation and control framework for training DRL agents \cite{jimenez2021}. \sinergym{} offers a standardized and flexible design that facilitates the comparison of algorithms under different environments, reward functions, states and actions spaces, as well as experiments replication.

The remainder of this paper is structured as follows. Section \ref{sec:2} details the theoretical background in which our research is framed. Section \ref{sec:3} introduces the main fundamentals of DRL and its application in HVAC control. Sections \ref{sec:4} and \ref{sec:5} will describe the environment and the experiments conducted, with the subsequent discussion on the results in section \ref{sec:6}. Finally, section \ref{sec:7} will detail the main conclusions derived from this research.


\section{Related work and novelty}
\label{sec:2}

The application of DRL in HVAC control is a developing field that has generated significant interest and growth within a broad community of researchers. Although the first approach to applying RL in building control dates back to 1998 by Mozer \cite{mozer1998}, the development of this field really began to be remarkable in the last few years, together with the rise of DRL. 

Extensive research about the application of DRL in building energy control and HVAC can be found in recent literature reviews, such as \cite{vazquez2019a,yu2021,wang2020b,han2019,leitao2020,mason2019,rajasekhar2020,zhang2018b,yang2020,mocanu2018}, where several applications, perspectives, and objectives within this field are summarized and studied in detail.

Focused on the DRL methods being used, we find many applications of well-established algorithms, such as Deep Q-Networks (DQN) \cite{lissa2021,yoon2019,gupta2021,sakuma2020}, Deep Deterministic Policy Gradient (DDPG) \cite{gao2020,zou2020}, or actor-critic methods \cite{morinibu2019,wang2017}. However, we share the opinion of Biemann et al. \cite{biemann2021} on the current situation in this field: (1) the widespread use of DQN ---instead of more modern algorithms such as Twin Delayed DDPG (TD3), Soft Actor-Critic (SAC) or Proximal Policy Optimization (PPO)--- in this continuous control problem could be hindering significant progress; and (2) these are generally isolated implementations that compare the performance of the chosen algorithm against a simplistic baseline and solve specific problems in well-delimited domains. In fact, there is little work benchmarking multiple DRL algorithms under standard conditions, as is common in other fields of machine learning \cite{vazquez2019a,wolfle2020,islam2017}.

One of the few exceptions is the aforementioned work by Biemann et al. \cite{biemann2021}, which develops a comparison between different DRL algorithms in a two-zone data center\footnote{\url{https://github.com/NREL/EnergyPlus/blob/develop/testfiles/2ZoneDataCenterHVAC_wEconomizer.idf}} environment. This is a well-known and commonly used test scenario in DRL-based HVAC control \cite{moriyama2018,zhang2019b}, in which the objective is the management of temperature setpoints and fan mass flow rates of a data center. Specifically, the algorithms they compared were SAC, TD3, PPO, and Trust Region Policy Optimization (TRPO), evaluating energy savings, thermal stability, robustness, and data efficiency. Their results revealed the predominance of off-policy algorithms and, specifically, SAC, achieving consumption reductions of approximately 15\% while guaranteeing comfort and data efficiency. Finally, the robustness was remarkable when the DRL algorithms were trained using different weather models, which increased their generalization and adaptability. 

Another remarkable and recent study is that of Brandi et al. \cite{brandi2022}, where the control of an HVAC system based on the charging and discharging of a refrigerated storage facility is presented. In this case, two RBCs and an MPC controller were compared with two SAC implementations: one trained offline ---in simulation---, and the other one online ---on the real building---. The authors obtained the best results for MPC and offline DRL, as well as an interesting approximation of the online trained SAC to the performance of its offline variant over time. A Long Short-Term Memory(LSTM) neural network architecture was used to predict the value of the variables composing the agents' observations over a given prediction horizon. The authors conclude that DRL-based solutions are more unstable than MPC solutions, although they exhibit adaptation to recurrent patterns without explicit supervision.

Regarding generalization capacity and robustness, few papers address this issue. In \cite{xu2020}, transfer learning \cite{torrey2010,taylor2009} is proposed to train and evaluate a DQN-based agent in a building layout, and then transfer it to a slightly different one. In this work, the controller uses two sub-networks: a front-end network, which captures building-agnostic behavior, and a back-end network, specifically trained for each building. This way, transfer learning avoids cold-starts and tuning in the second building, allowing for rapid deployment and greater efficiency. Different experiments are carried out, such as (1) transfer from $n$-zone to $n$-zone buildings with different materials and layouts; (2) transfer from $n$-zone to $m$-zone; and (3) transfer from $n$-zone to $n$-zone with different HVAC equipment. A second related work is \cite{lissa2021}, which focuses on spatial and geographical variations in HVAC control systems. Among their main achievements, we can find a significant reduction in the time required to reach comfort temperatures, although the authors do not delve into energy efficiency.

At this point, we can identify certain limitations and research gaps in the relevant but scarce literature addressing the experimental evaluation of DRL algorithms in HVAC control:

\begin{itemize}
    \item Studies such as \cite{biemann2021} and \cite{brandi2022} only provide results in a single environment. In contrast, benchmarking is desirable to be performed using a representative enough set of scenarios and test configurations.
    \item Something similar happens when it comes to the application of transfer learning: the existing case studies are limited and use different environments, so they do not enable comparisons to be made. We believe that further deepening of its application is necessary, concretely by promoting, as far as possible, the generalization capacity and robustness of different agents in a broader set of testing environments.
    \item There are additional opportunities close to transfer learning that have not yet been addressed, such as the application of sequential learning. The objective pursued is to compare if a controller that learns progressively to manage an HVAC system under different weather conditions can be more efficient than one directly trained on a fixed setting.
    \item Finally, we consider it relevant to study how the manipulation of the reward function used by DRL agents affects their performance, and what consequences these changes may have in terms of consumption and comfort violation.
\end{itemize}


\section{Theoretical background}
\label{sec:3}

\subsection{Deep Reinforcement Learning}
\label{sec:3.1}

A reinforcement learning problem is defined as a Markov Decision Process (MDP) consisting of the following elements:

\begin{itemize}
    \item An agent, which learns from the interaction with its environment over a discrete sequence of time steps $\mathcal{T} = \{0,1,2,...\}$ and that pursues a certain objective.
    \item An environment, defined as any element external to the agent. It is a dynamic process that produces relevant data for the agent. A state $s_t \in \mathcal{S}$ represents the current situation of the environment at time step $t$.
    \item A set of actions $\mathcal{A}$, which determine the dynamics of the environment. Every action $a_t \in \mathcal{A}$ allows the agent to transit between different states.
    \item A reward function that evaluates the goodness of an action or state for the agent. The reward signal $r_t \in \mathbb{R}$ drives the agent's training.
    \item A policy function $\pi : \mathcal{S} \rightarrow \mathcal{A}$, which determines the actions to be taken by the agent in the face of different states.
\end{itemize}

The goal of an RL agent is to achieve optimal behavior towards accomplishing a task, which involves learning an optimal policy that maximizes cumulative reward \cite{zhang2020, sutton2018}. Learning this optimal policy can be solved through dynamic programming in problems with not-particularly-large action and/or state spaces. For more complex cases, there are a variety of algorithms based on iterative improvement of the agent's policy or successive approximation of the expected reward of states and actions, e.g. Monte Carlo, SARSA, or the widely used Q-learning \cite{watkins1992}.


However, these methods generally become infeasible as the complexity of the environment increases. This may be due to considerable growth in the number of actions or states that compose the environment, which can be mitigated by ad hoc methods that are difficult to generalize. Here is where deep learning comes into play, as neural networks can be applied to learn abstract representations of states and actions, as well as to approximate the expected rewards based on historical data. The combination of RL with deep neural networks has given rise to Deep Reinforcement Learning (DRL), which has lead to numerous applications of reinforcement learning in real environments. Thus, DRL algorithms combine the best of both worlds, endowing reinforcement learning with the representational power, efficiency, and flexibility of deep learning \cite{zai2020}. 

In recent years, we have observed several advances in the development of DRL algorithms that represent the state of the art, such as Deep Q-Networks (DQN) \cite{mnih2013}, Deep Deterministic Policy Gradient (DDPG) \cite{lillicrap2015},  Asynchronous Advantage Actor-Critic (A3C) and Advantage Actor-Critic (A2C) \cite{mnih2016}, Trust Region Policy Optimization (TRPO) \cite{schulman2017b}, Proximal Policy Optimization (PPO) \cite{schulman2017a}, Twin Delayed DDPG (TD3) \cite{fujimoto2018}, and Soft Actor-Critic (SAC) \cite{haarnoja2018}. \edited{Table \ref{tab:drl-algorithms} summarises the main features and limitations of these algorithms.}

\begin{table}
    \caption{\edited{Some of the most widely used DRL algorithms, including their main features and limitations}}
    \label{tab:drl-algorithms}
    \centering
    \resizebox{\textwidth}{!}{%
    \begin{tabular}{ll}
    \hline
    \textbf{Algorithm} & \textbf{Main features and limitations}                                                                                                                                                                                                                           \\ \hline
    \textbf{DQN}       & \begin{tabular}[c]{@{}l@{}}+ Simple value-based off-policy algorithm.\\ + Implements a replay buffer and a target network.\\ -- Limited to discrete action spaces and prone to overestimating action values.\\ -- Improved by variants like DDPG.\end{tabular} \\ \hline
    \textbf{DDPG}      & \begin{tabular}[c]{@{}l@{}}+ Off-policy algorithm. \\ + Improves DQN by allowing continuous action spaces.\\ -- Highly sensitive to hyperparameterisation. \end{tabular}                                                                                     \\ \hline
    \textbf{A2C/A3C}   & \begin{tabular}[c]{@{}l@{}}+ On-policy gradient-based algorithms focused on parallelisation.\\+ Based on the joint learning of multiple agents implementing local policies.\\+ A2C includes a coordinator to synchronise all agents.\\ -- Underperforms more complex policy-gradient methods.\end{tabular}                                \\ \hline
    \textbf{TRPO}      & \begin{tabular}[c]{@{}l@{}}+ On-policy gradient-based algorithm.\\+ Employs a trust region defined by a KL-divergence constraint to limit policy updates.\\ -- More complex and less stable than PPO.\end{tabular}                               \\ \hline
    \textbf{PPO}       & \begin{tabular}[c]{@{}l@{}}+ On-policy algorithm based on policy gradient.\\+ Improves TRPO by adding a clipped objective to prevent large policy updates.\\+ Reduced computational costs.\\ -- Data-inefficient. \end{tabular}                                       \\ \hline
    \textbf{TD3}       & \begin{tabular}[c]{@{}l@{}}+ Off-policy actor-critic algorithm.\\ + Extends DDPG with improved stability and convergence guarantees. \\ + Clipped double-Q learning, delayed policy updates and target policy smoothing. \\ -- May experience action saturation.\end{tabular}  \\ \hline
    \textbf{SAC}       & \begin{tabular}[c]{@{}l@{}}+ Off-policy actor-critic algorithm.\\ + Employs entropy regularization to promote stability and exploration. \\ + Less sensible to hyperparameterisation.\\ -- Slow training.\end{tabular}                                                 \\ \hline
    \end{tabular}%
    }
\end{table}

As we will see in the following subsection, some of these algorithms are powerful tools for solving continuous control problems with an indefinite number of states and actions \cite{duan2016}, as is the case of HVAC control.

\subsection{HVAC control problem formulation}
\label{sec:3.2}

The feasibility of HVAC control through DRL was firstly demonstrated by using reduced action spaces and simplified building models \cite{zhang2019a,moriyama2018}. The objective of this problem is to find an optimal control policy that maximises the comfort of occupants while minimising energy consumption. The problem is formulated as a Partially Observable MDP (POMDP), and the goal is to find a policy that includes as many desirable states as possible while avoiding energetically-costly actions. Additional parameters, such as energy price and the use of renewable energy sources, can also be considered in the optimization process. However, for the sake of simplicity, we will focus on comfort and consumption as the primary targets.

Based on a set of observed variables that define the ambient conditions of the environment ---i.e., outdoor/indoor temperature, CO$_2$ concentration, humidity, occupancy---, the following objectives are considered:

\begin{itemize}
    \item \edited{Regarding power demand $P$, we seek a policy that leads to its minimization:}
    
        \begin{equation}
            \pi^* = \underset{\pi_\theta}{argmin}\sum^T_{t=1}\ P_t
        \end{equation}
     
    \item \edited{In the case of comfort, we try to minimize the distance $C$ between the current state of the building, $S_t$, and a target state, $S_{target}$:}
    
        \begin{equation}
            \pi^* = \underset{\pi_\theta}{argmin}\sum^T_{t=1}\ C(S_t, S_{target})
        \end{equation}
\end{itemize}

A state will be desirable if the variables that compose it are within certain pre-established preferences. Therefore, we say that a comfort violation occurs if the value of these variables differs from the established limits.

\edited{In this problem, $P$ corresponds to the consumption of the building's thermal equipment ---e.g., heat pump, boiler--- while $C(S_t, S_{target})$ refers to the difference between the actual and desired temperatures in the controlled zones of the building.}

Therefore, we can combine the minimization of power demand and the maximization of comfort in a single expression:

\begin{equation}
    \pi^* = \underset{\pi_\theta}{argmin}\sum^T_{t=1} \omega \ C(S_t,S_{target}) + (1 - \omega) \ P_t
\end{equation}

with $\omega$ and $(1-\omega)$ being the weights assigned to comfort and power demand, respectively. We can now define our reward function such that:

\begin{equation}
    r(S_t, A_t) = (1 - \omega) \ \lambda_P \ P_t + \omega \ \lambda_C \ C(S_t, S_{target})    
\end{equation}

\edited{where $P_t$ is the power demand of the HVAC system at time step $t$; $C(S_t, S_{target})$ represents the distance from the current zone temperature to the desired comfort limits}; $\omega$ represent the weighting assigned to each part of the reward, and $\lambda_P$ and $\lambda_C$, two constant factors used for scaling the magnitudes of power demand and temperature.

It is common to express this reward in negative terms, since its maximization is sought, which leads us to rewrite it as:

\begin{equation}
    \label{eq:reward}
    r(S_t, A_t) = -(1 - \omega) \ \lambda_P \ P_t - \omega \ \lambda_C \ C(S_t, S_{target})   
\end{equation}

Note how the reward function employed will vary depending on the variables to be considered in the problem. As discussed later, the details of this function is one of the main design decisions in DRL.
  
Finally, the actions to be performed by the agent will depend on the problem we are facing with. For example, in problems where the actions consist of adjusting the heating and cooling setpoints of the HVAC system, the action space may be discrete ---there is a finite number of actions where each action is a tuple with fixed setpoint values---, or continuous ---each setpoint is a real number to be adjusted---. Other similar problems can be addressed with a similar setup, such as air flow regulation \cite{raman2020} or lighting control \cite{chen2018}.

\subsection{Simulation models}
\label{sec:3.3}

Most of the DRL solutions proposed in the literature are not directly trained on a real building, but they make use of different simulation tools. In the HVAC context, \href{EnergyPlus}{https://energyplus.net/} and \href{Modelica}{https://modelica.org/} are the most widely used frameworks. The use of this type of simulators is motivated by the fact that training a DRL agent in a real scenario would be too inefficient, as there is a need to establish a complete mapping between states, actions, and rewards, also considering extreme cases \cite{brandi2020}. In fact, it has been observed that it takes 20-50 days of training to converge on an acceptable control policy \cite{fazenda2014, costanzo2016, vazquez2019b}, thus making impractical to directly train DRL agents while they are deployed.

Typically, simulators are combined with deep learning (e.g., TensorFlow, Keras, PyTorch) or DRL libraries (e.g., Stable Baselines, RLlib, TensorFlow Agents) to pre-train and test algorithms in simulated environments prior to deployment \cite{valladares2019,vazquez2019b}. However, communication between DRL agents and simulation engines is rarely straightforward, which leads us to tools such as \boptest{} \cite{blum2021}, \energym{} \cite{scharnhorst2021}, \rltestbed{} \cite{moriyama2018} or \sinergym{}. These tools enable the communication between DRL agents and energy simulators, providing the necessary tools to perform the training and evaluation of the algorithms in different settings.

In the case of \sinergym{}, this software acts as a wrapper for EnergyPlus, offering several tools that enable DRL agents execution, data logging, the configuration of simulation environments, as well as the customization of states, actions and rewards. Its similarities and distinctive features with respect to other alternatives are detailed in Table \ref{tab:frameworks}.

\begin{table}
    \caption{\edited{Frameworks for DRL-based building energy optimization. \textbf{Custom spaces}: observation and actions spaces customization. \textbf{Actions}: support for continuous and discrete action spaces. \textbf{Weather config.}: weather customization capabilities. \textbf{Custom rewards}: support for customized reward functions. \textbf{Simulator}: building simulation engine. \textbf{Envs.}: number of environments available. Adapted and updated from \cite{jimenez2021}}}
    \label{tab:frameworks}
    \centering
    \resizebox{\textwidth}{!}{%
    \begin{tabular}{lllllll}
    \hline
    \textbf{Framework}                                                            & \textbf{\begin{tabular}[c]{@{}l@{}}Custom\\ spaces\end{tabular}} & \textbf{\begin{tabular}[c]{@{}l@{}}Actions\end{tabular}}  & \textbf{\begin{tabular}[c]{@{}l@{}}Weather\\ config.\end{tabular}} & \textbf{\begin{tabular}[c]{@{}l@{}}Custom\\ rewards\end{tabular}} & \textbf{Simulator}                                                       & \textbf{Envs.} \\ \hline
    \begin{tabular}[c]{@{}l@{}}RL TestBed for\\ EnergyPlus \cite{moriyama2018} \end{tabular}  & \xmark                                                                               & Continuous                                                        & \xmark                                                                       & \xmark                                                                & EnergyPlus 9.5.0                                                         & 3                     \\ \hline
    Energym \cite{scharnhorst2021}                                                             & \xmark                                                                               & Continuous                                                        & \cmark                                                                   & \cmark                                                               & \begin{tabular}[c]{@{}l@{}}EnergyPlus 9.5.0\\ Modelica 2.14\end{tabular} & 14                    \\ \hline
    BOPTEST \cite{blum2021}                                                              & \cmark                                                                                & \begin{tabular}[c]{@{}l@{}}Discrete,\\ continuous\end{tabular} & \xmark                                                                       & \cmark                                                                 & Modelica 4.0.0                                                           & 7                     \\ \hline
    \textbf{Sinergym} \cite{jimenez2021}                                                          & \cmark                                                                                & \begin{tabular}[c]{@{}l@{}}Discrete,\\ continuous\end{tabular} & \cmark                                                                        & \cmark                                                                 & EnergyPlus 23.1.0                                                        & 87                    \\ \hline
    \end{tabular}%
    }
\end{table}


\section{Data and environments}
\label{sec:4}

All the experiments were formulated and executed using the \sinergym{} framework \cite{jimenez2021}. In the following subsections we present the configuration details, including state and action spaces, reward functions and weather data.

\subsection{Building models}

\begin{figure}
    \centering
        \begin{subfigure}{0.7\textwidth}
            \includegraphics[width=\textwidth]{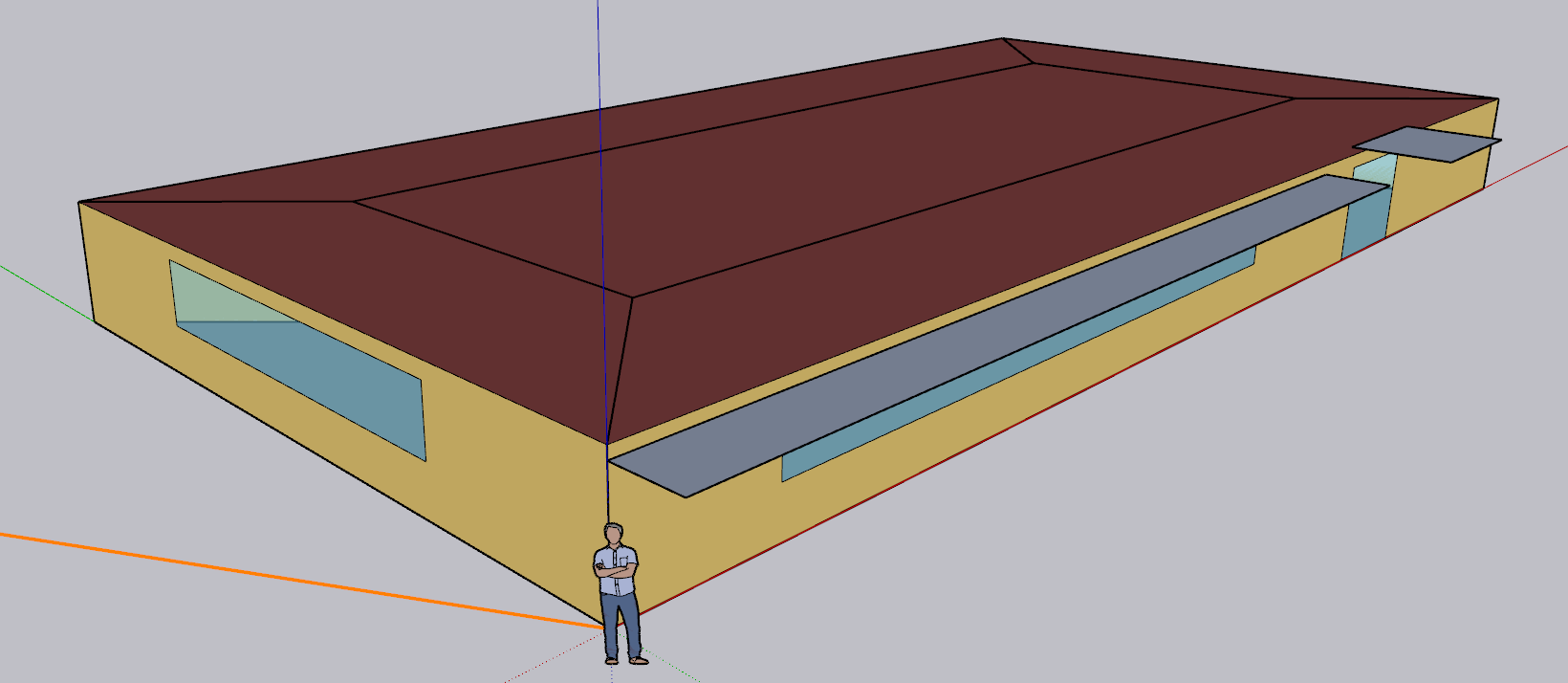}
            \caption{\fivezone{} representation}
            \label{fig:5zone-idf}
        \end{subfigure}
        \begin{subfigure}{0.7\textwidth}
            \includegraphics[width=\textwidth]{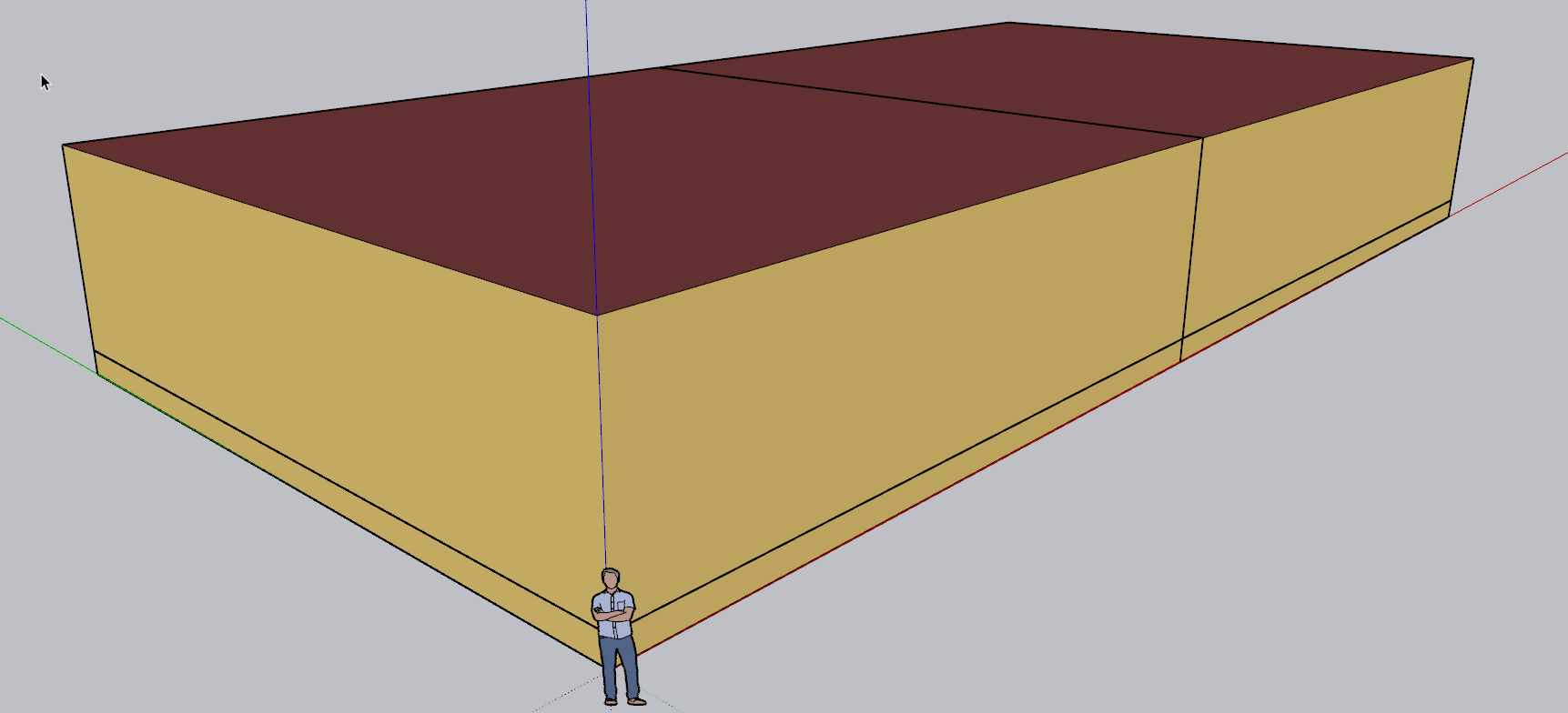}
            \caption{\datacenter{} representation}
            \label{fig:datacenter-idf}
        \end{subfigure}   
    \caption{Building models used as benchmark environments}
    \label{fig:buildings}
\end{figure}

The following building models included in \sinergym{}, were chosen as test scenarios:

    \begin{itemize}
        \item \textbf{\fivezone{}} \cite{wei2017,ding2020}. This is a single-story rectangular office building with dimensions of $30.48 m \times 15.24 m$ (see Figure \ref{fig:5zone-idf}). The building is divided into 5 zones: 4 exterior zones and 1 interior zone. The average height of the building is $3.048 m$ and the total surface area is $463.6 m^2$. it is equipped with a packaged \edited{Variable Air Volume} (VAV), \edited{Direct Expansion} (DX) cooling coil and gas heating coils, with fully auto-sized input as the HVAC system to be controlled.
        \item \textbf{\datacenter{}} \cite{moriyama2018,zhang2019b,li2019,biemann2021}. A 1-story data center with a surface of $491.3 m^2$ (see Figure \ref{fig:datacenter-idf}). The building is divided into two asymmetrical zones (east and west). Each data center zone has an HVAC system consisting on an air economizer, direct and indirect evaporative coolers, a single speed DX cooling coil, chilled water coil, and VAV with no reheat air terminal unit.
    \end{itemize}

In both cases, we aim to regulate indoor temperatures by balancing comfort and power demand while also considering the influence of one on the other.

\subsection{Weather}

Given the importance of weather conditions in building control strategies \cite{ghahramani2016}, the experimentation was conducted considering three types of weathers integrated in \sinergym{}, based on Typical Meteorological Year (TMY3) data obtained from \cite{energycodes2021}. Note how these climates vary in their average temperatures and humidity levels, thus providing a diverse and representative test set. These are:

\begin{itemize}
    \item \textbf{Hot dry}. Climate corresponding to Arizona (USA), with an average annual temperature of $21.7$\textcelsius~and an average annual relative humidity of 34.9\%.
    \item \textbf{Mixed humid}. Climate in New York (USA), with an average annual temperature of 12.6\textcelsius~and an average annual relative humidity of 68.5\%.
    \item \textbf{Cool marine}. Corresponds to Washington (USA). The average annual temperature is 9.3\textcelsius, while the average annual relative humidity is 81.1\%.
\end{itemize}

\sinergym{} offers the possibility of adding random variations to the weather data between episodes by using Ornstein-Uhlenbeck (OU) processes \cite{benth2005,jimenez2021}. This enables a more varied training aimed at preventing overfitting and improving generalization. Thus, in this work, stochasticity was achieved by applying OU with the following parameters: $\sigma=1.0$, $\mu=0$ and $\tau=0.001$.

\subsection{Observation and action spaces}
\label{sec:spaces}


At each simulation step, the agent receives information from the environment in the form of an observation $o_t \in \mathcal{O} \subset \mathcal{S}$. The information  received consists of a subset of variables that provide information about the current state of the building, including comfort and consumption metrics, occupancy, time information, indoor and outdoor temperatures, humidity, wind, or solar radiation. These variables may vary in number or zone depending on the building we are trying to control. However, EnergyPlus uses common identifiers shared by different buildings, which facilitates the management of this information.

Therefore, Table \ref{tab:observation-variables} contain the variables that make up an observation for \fivezone{} and \datacenter{}, as well as their ranges\footnote{For an in-depth study of these variables, please refer to EnergyPlus official documentation: \url{https://energyplus.net/documentation}}. Both the observed and controlled variables are extracted from the simulation of each building, with 20 for \fivezone{} and 29 for \datacenter{}. \edited{Depending on the building and simulation conditions, each variable can take values in a wide range that we are unable to determine a priori. Therefore, most of the variables are assumed to take values in $[-5e6, 5e6]$, as these are the minimum and maximum limits of the simulator outputs. However, all the observation values were normalized in $[0,1]$ using the normalization wrapper provided by \sinergym{}.}

\begin{table}
    \footnotesize
    \caption{\edited{Observation variables in \fivezone{} and \datacenter{}. In \datacenter{}, variables marked with an asterisk are recorded for both east and west zones}}
    \label{tab:observation-variables}
    \centering
    \begin{tabular}{lll}
    \hline
    \textbf{Variable name}                          & \textbf{Min.} & \textbf{Max.} \\ \hline
    Site Outdoor Air Drybulb Temperature            & -5e6          & 5e6           \\
    Site Outdoor Air Relative Humidity              & -5e6          & 5e6           \\
    Site Wind Speed                                 & -5e6          & 5e6           \\
    Site Wind Direction                             & -5e6          & 5e6           \\
    Site Diffuse Solar Radiation Rate per Area      & -5e6          & 5e6           \\
    Site Direct Solar Radiation Rate per Area       & -5e6          & 5e6           \\
    Zone Thermostat Heating Setpoint Temperature *  & -5e6          & 5e6           \\
    Zone Thermostat Cooling Setpoint Temperature *  & -5e6          & 5e6           \\
    Zone Air Temperature *                          & -5e6          & 5e6           \\
    Zone Thermal Comfort Mean Radiant Temperature * & -5e6          & 5e6           \\
    Zone Air Relative Humidity *                    & -5e6          & 5e6           \\
    Zone Thermal Comfort Clothing Value *           & -5e6          & 5e6           \\
    Zone Thermal Comfort Fanger Model PPD *         & -5e6          & 5e6           \\
    Zone People Occupant Count *                    & -5e6          & 5e6           \\
    People Air Temperature *                        & -5e6          & 5e6           \\
    Facility Total HVAC Electricity Demand Rate     & -5e6          & 5e6           \\ \hline
    Current year                                    & 1             & 2050          \\
    Current month                                   & 1             & 31            \\
    Current day                                     & 1             & 12            \\
    Current hour                                    & 0             & 23            \\ \hline
    \end{tabular}
\end{table}

Note that a comprehensive analysis of the variables required for training will depend on the problem's context and the scope of the controller's application. In this case, to train all agents with the most information available, we did not explicitly conduct a detailed study of these observations.


Regarding the action space, each action consists of a set of values representing the HVAC heating and cooling setpoints. In the case of \fivezone{}, an action $a_t \in \mathbb{R}^2$ involves the selection of two values representing the heating and cooling setpoints for the whole building. Conversely, for \datacenter{}, an action $a_t \in \mathbb{R}^4$ consists of four values corresponding to the selected setpoints for the east and west zones of the facility. Their identifier, definition, and ranges are displayed in Table \ref{tab:action-variables}.

\begin{table}
    \footnotesize
    \caption{\edited{Action variables for \fivezone{} and \datacenter{}}}
    \label{tab:action-variables}
    \begin{tabular}{llll}
        \hline
        \textbf{Building}        & \textbf{Variable name}                 & \textbf{Min.} & \textbf{Max.} \\ \hline
        \fivezone{}              & Zone Heating Setpoint Temperature      & 15            & 22.5          \\
        \multicolumn{1}{c}{}     & Zone Cooling Setpoint Temperature      & 22.5          & 30            \\ \hline
        \datacenter{}            & East Zone Heating Setpoint Temperature & 15            & 22.5          \\
                                 & East Zone Cooling Setpoint Temperature & 22.5          & 30            \\
                                 & West Zone Heating Setpoint Temperature & 15            & 22.5          \\
                                 & West Zone Cooling Setpoint Temperature & 22.5          & 30            \\ \hline
        \end{tabular}
\end{table}

Finally, based on several tests, we consider a continuous action space, as opposed to a discrete one. This is because a discrete action space can overly constrain the agent's behaviour if the number of allowed actions is limited. In addition, the use of fixed setpoints can worsen the agent's flexibility by explicitly imposing expert knowledge. In our case, while expert knowledge is present in the range of values that these variables can take, there are no restrictions within this range, allowing each setpoint to take arbitrary continuous values ---as long as the heating setpoint remains below the cooling setpoint---. Finally, we must consider that a continuous action space covers the action range of a discrete one. This means that the possibility of choosing similar actions is not lost, while offering a wide margin for improvement.

\subsection{Reward}

For the reward calculation, we chose the linear reward function included by default in \sinergym{}:

\begin{equation}
\label{eq:linear-reward}
     r_t = - \omega \ \lambda_P \ P_t - (1 - \omega) \ \lambda_C \ (\mid T_t - T_{up} \mid_+ + \mid T_t - T_{low} \mid_+)
\end{equation}

This is a similar formula to the one shown in Equation \ref{eq:reward}, where $\omega$ is the weight assigned to power consumption ---and consequently, $1 - \omega$ the weight assigned to comfort---; $\lambda_P$ and $\lambda_C$ are both scaling constants; $P_t$ is the facility total HVAC power demand rate (measured in W); $T_t$ is the current inner temperature (in \textcelsius), and $T_{up}$ and $T_{low}$ are the limits of the target comfort range. Thus, the greater the consumption, or distance from the interior of the comfort range, the worse the reward, and the more the agent is penalized.

\edited{The linear reward allows us to maintain a real trade-off between comfort and consumption. Thus, assuming precise scaling factors, the preference for each reward term is determined solely by the value of $\omega$, and no term is more important than the other by default.}

\edited{Recent studies such as \cite{kadamala2024enhancing} address the issue of reward selection without reaching a clear consensus on which option is more effective in practice, so we opt for the most equitable reward definition. Other alternatives may lead to a bias towards a particular reward term. For example, this is the case with \sinergym{}'s exponential reward, where deviations from target temperatures have a greater impact than increases in power consumption.}

For the majority of the experiments in section~\ref{sec:5}, we use the default value of $\omega = 0.5$, which involves seeking a complete balance between comfort and power requirements. We also analyse the effect of using different weightings in section~\ref{sec:5.5}.

\subsection{Algorithms}

We briefly summarize the DRL algorithms and rule-based controllers used throughout the experimentation. \edited{The proposed control problem is addressed from a single-agent perspective, since multi-agent HVAC control is mostly applied in building communities \cite{pinto2021coordinated,vazquez2020marlisa} or problems where the action space is too large for a single agent \cite{fu2022optimal}. For the environments considered in this work ---with a maximum of two controlled zones in \datacenter{}---, such control is not strictly required}.

The DRL algorithms selected were PPO, TD3, and SAC. As it is beyond the scope of this paper to go into their theoretical details, we refer the reader to the original publications where these are presented in detail: \cite{schulman2017a} (PPO), \cite{fujimoto2018} (TD3), and \cite{haarnoja2018} (SAC).

PPO is an improvement over the on-policy algorithm TRPO. On-policy algorithms have a single policy that improves progressively according to the exploration of states and/or actions. This algorithms are generally sample-inefficient due to the data loss that occurs when updating their policy. This may hinder their performance in domains such as HVAC control, where data collection is slow, as detailed in \cite{biemann2021}. This is an assumption that we will empirically test in our experiments.

On the other hand, we have off-policy algorithms such as TD3 and SAC, which are more sample-efficient and can therefore be expected to produce better results in the HVAC control problem. Off-policy algorithms are based on two policies: a behaviour policy, which is exploratory, and a target policy, which is the one actually used by the agent and is based on the knowledge gathered by the exploratory one. Thus, while TD3 is an improved version of DDPG commonly used in HVAC control \cite{li2019, gao2020,biemann2021,fu2021}, SAC offers a promising alternative in this domain, as shown by recent studies such as \cite{brandi2022,yu2020,biemann2021,coraci2021}.

Regarding the rule-based controllers used as baselines, the following approaches produced the best results for each building:

\begin{itemize}
    \item For \fivezone{}, a control based on static setpoints depending on the season of the year, based on the ASHRAE standard \cite{ashrae2004} for thermal comfort in dwellings. In this way, the static setpoints $[26, 29]$ \textcelsius~are set for the period from June to September, while setpoints $[20, 23.5]$ \textcelsius~are maintained for the rest of the year.
    \item In the case of \datacenter{}, an integral control based on degree-by-degree correction of the current setpoints, depending on whether the temperature of each zone is within the specified range. This range corresponds to ASHRAE standard \cite{ashrae2016} for recommended temperature ranges in data centers, which establishes $[18, 27]$ \textcelsius~as reference setpoint values in these facilities.
\end{itemize}

Although these RBCs are able to guarantee proper control in the proposed scenarios, it should be noted that (1) they are not easily scalable, as they have to be defined manually; (2) they are not able to consider wide optimization horizons, as the rules used are essentially reactive; and (3) in more complex scenarios (e.g. larger number of variables to be controlled), their performance could be significantly compromised.

For an in-depth study on how these RBCs are implemented, we refer the reader to the corresponding \sinergym{} \textit{\href{controllers}{https://github.com/ugr-sail/sinergym/blob/main/sinergym/utils/controllers.py}} module, where their source code can be consulted.


\section{Experiments}
\label{sec:5}

The following subsections detail the different experiments conducted. The objective pursued was to thoroughly study (1) which algorithms are able to guarantee better thermal control; (2) which DRL algorithms offer higher robustness to weather conditions for which they have not been trained; (3) the application of sequential learning and its effectiveness to obtain better controllers, and (4) the comfort-consumption trade-off and its influence on the performance of the controllers.

Although the most significant results will be illustrated graphically, space limitations have led us to transfer part of the complementary graphic material to an external repository: \url{https://github.com/ugr-sail/paper-drl_building}, available for detailed analysis of simulation data, code, and additional charts.

\subsection{Methodology}
\label{sec:5.0}

\begin{figure}
    \centering
    \includegraphics[width=\textwidth]{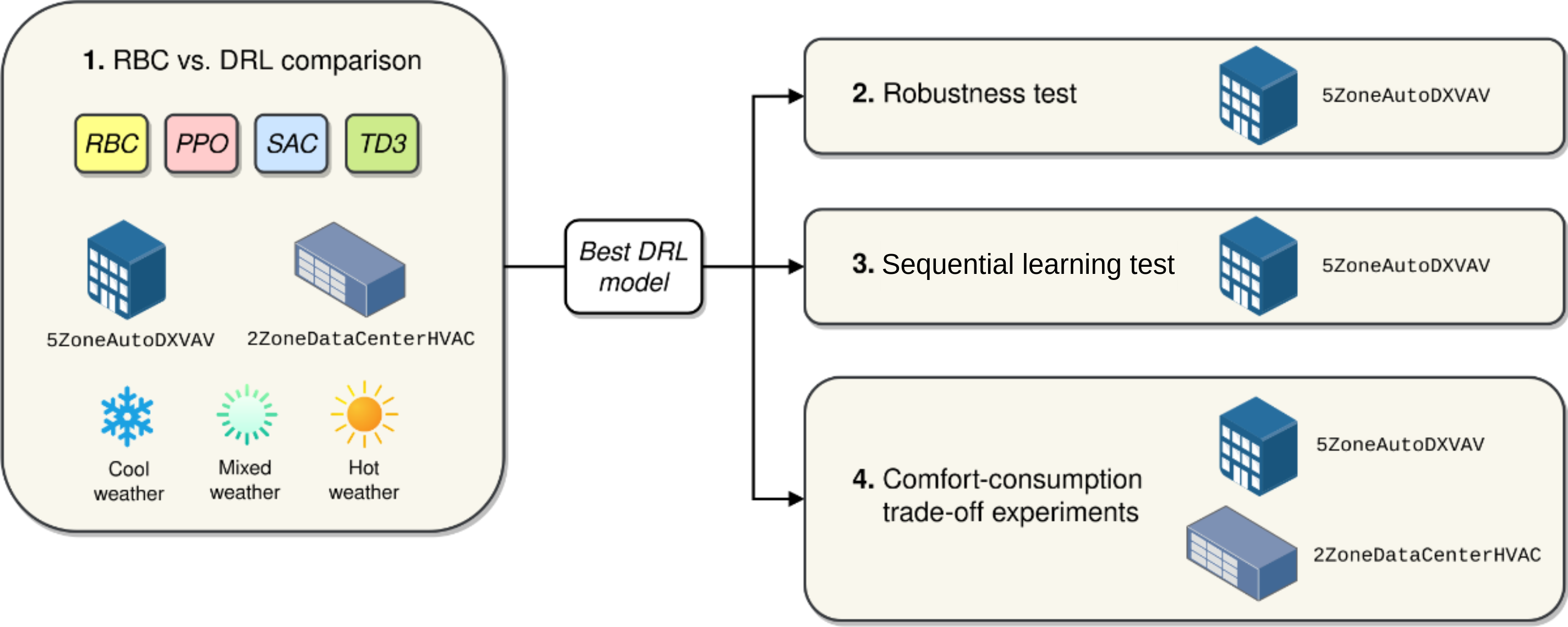}
    \caption{Proposed experiments}
    \label{fig:methdology}
\end{figure}

Figure~\ref{fig:methdology} summarizes the experimentation phases followed in this work. First, the performance between RBCs and DRL algorithms were compared in the \fivezone{} and \datacenter{} buildings with three different types of weather conditions, that is, 6 different environments.

Following this comparison, further experiments were carried out under more complex settings. On the one hand, the best DRL agent obtained in the previous step was used to test how it adapted to different weather conditions than those used during its training. We refer to this experiment as the ``robustness test'', which allowed us to evaluate the agent's ability to generalise to situations not experienced during training.

Subsequently, we explored the application of sequential learning by progressively training an agent over multiple weather conditions, comparing its performance with that of an agent specialised in a single environment

Finally, we compared the performance of the best DRL controllers of each building under different definitions of the reward function, mainly by varying the weights of the comfort and consumption terms and observing the results obtained.

\subsection{Experimental settings}
\label{sec:5.1}

Both the training and evaluation of the DRL algorithms were executed using Google Cloud computing platform and \sinergym{} version \href{1.8.2}{https://pypi.org/project/sinergym/1.8.2/}. Regarding the handling of simulation time, each episode corresponds to 1 year of building simulation. Each episode consists of 35040 time steps (i.e., 365 days $\times$ 24 hours/day $\times$ 4 timesteps/hour), resulting in an observation sample and subsequent setpoint adjustment every 15 minutes. This is an appropriate value for the problem addressed, although it can be configured as desired within \sinergym{} settings.

In addition, the evaluation metrics common to all experiments and used to compare the agents are the following:

\begin{itemize}
    \item \textbf{Mean episode reward}: calculated as the arithmetic mean of the rewards obtained in each time step of an episode.
    \item \textbf{Mean power demand}: mean HVAC electricity demand rate of the building's HVAC system (in W), provided by EnergyPlus.
    \item \textbf{Comfort violation}: percentage of time during which the ambient temperature is outside a desired comfort range, and mean value (in \textcelsius) of temperature deviations from comfort limits. In the case of \fivezone{}, the comfort ranges used were: $[23, 26]$ \textcelsius{} for June to September, and $[20, 23.5]$ \textcelsius{} for the rest of the year. Meanwhile, for \datacenter{}, a single comfort range was used: $[18, 27]$ \textcelsius, as recommended by ASHRAE's standard \cite{ashrae2016}, in order to ensure the safety of the building's equipment.
\end{itemize}

Other metrics, such as the evolution of indoor and outdoor temperatures, will also give an insight into how the control adjusts between setpoints and how performance evolves, as detailed below.

\subsection{Comparison between control algorithms}
\label{sec:5.2}

The first experiment conducted was a comparison between DRL algorithms and RBCs. This involve the training and subsequent evaluation of the DRL algorithms in different environments, in order to identify which algorithms perform better for each environment. The environments used for training and evaluation consisted in all the combinations between buildings (\fivezone{}, \datacenter{}) and weathers (hot dry, mixed humid, and cool marine), resulting in 6 different scenarios.

The configuration settings for this experiment were as follows:

\begin{itemize}
    \item \textbf{Machine used for training}: Google Cloud's \href{n2-highcpu-8}{https://cloud.google.com/compute/docs/machine-types} machine, equipped with a 2.2 GHz \textit{Intel Cascade Lake} processor, 8 vCPU, and 8192 MiB of RAM.
    \item \textbf{Number of training episodes}: 20. This was found to be a sufficient number of episodes for all algorithms to achieve convergence.
    \item \textbf{Frequency of evaluation}: 4. Specifies the number of training episodes after which an evaluation and selection of the best model is performed.
    \item \textbf{Evaluation length}: 3. Refers to the number of episodes used to perform the evaluations.
    \item \textbf{Random seed}: 42. This seed was used in all experiments to facilitate the replication of the results.
    \item \textbf{Normalization} of the observations in $[0, 1]$, using the \sinergym{} normalization wrapper.
    \item Multiple combinations of \textbf{hyperparameters} were tested until finding the ones that offered the best results for each DRL algorithm and environment combination, as detailed in Appendix \ref{sec:appendixA}. Endorsing the idea of \cite{brandi2020, agarwal2021}, the choice of hyperparameters in DRL is a complex task that greatly affects the training process of the agents. Therefore, it is necessary to follow a testing process during several episodes to compare the performance of different configurations. This philosophy was followed until acceptable sets of hyperparameters were found.
\end{itemize}

\begin{figure}
    \centering
    \includegraphics[width=\linewidth]{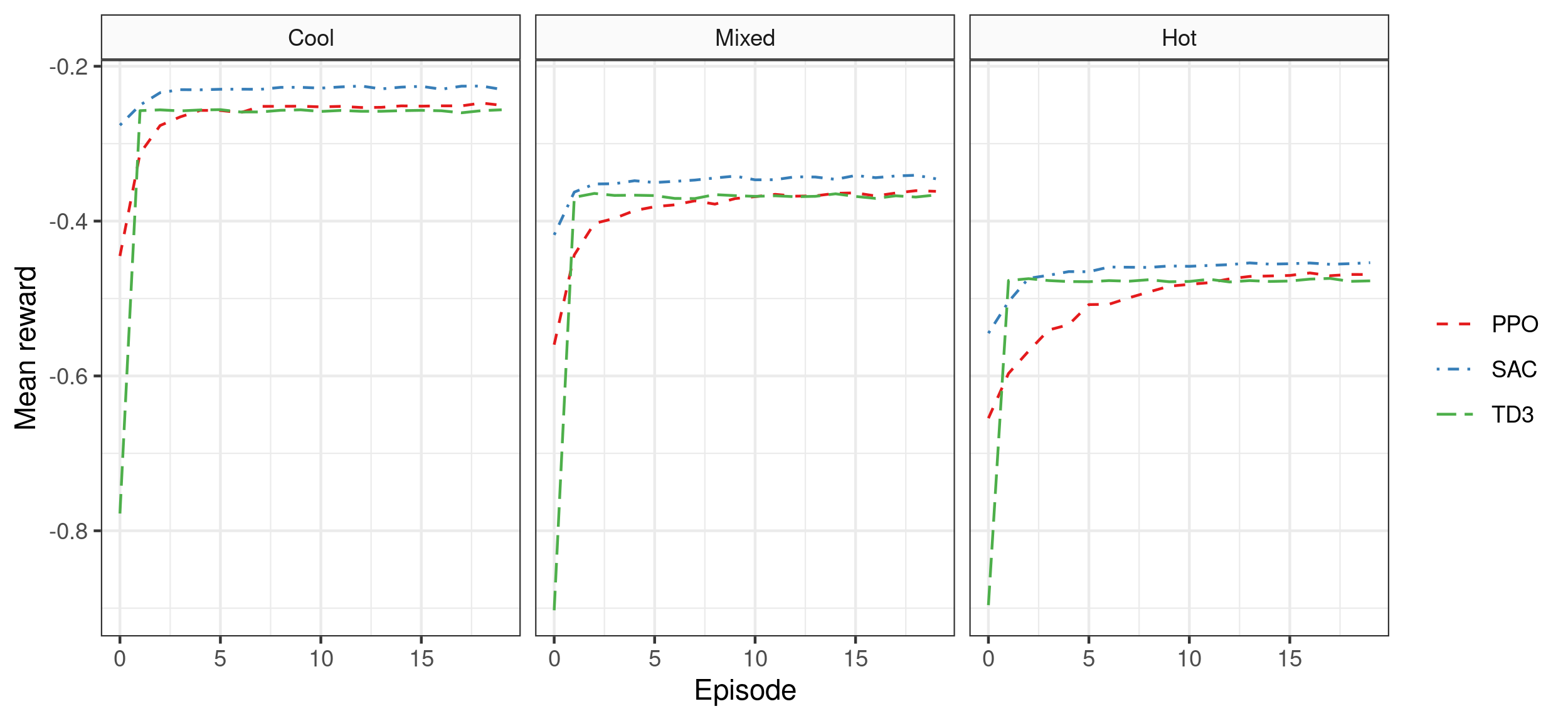}
    \caption{Mean episode rewards during 20 training episodes in \fivezone{}}
    \label{fig:train_rewards_5Zone}
\end{figure}

Regarding the training process, Figure \ref{fig:train_rewards_5Zone} shows how SAC offers the best results in all \fivezone{} scenarios, with early convergence at its optimal values. Moreover, TD3 shows fast convergence, although without surpassing SAC, and offering similar results ---or even slightly lower--- than those of PPO, whose convergence requires a greater number of training episodes.

\begin{figure}
    \centering
    \includegraphics[width=\linewidth]{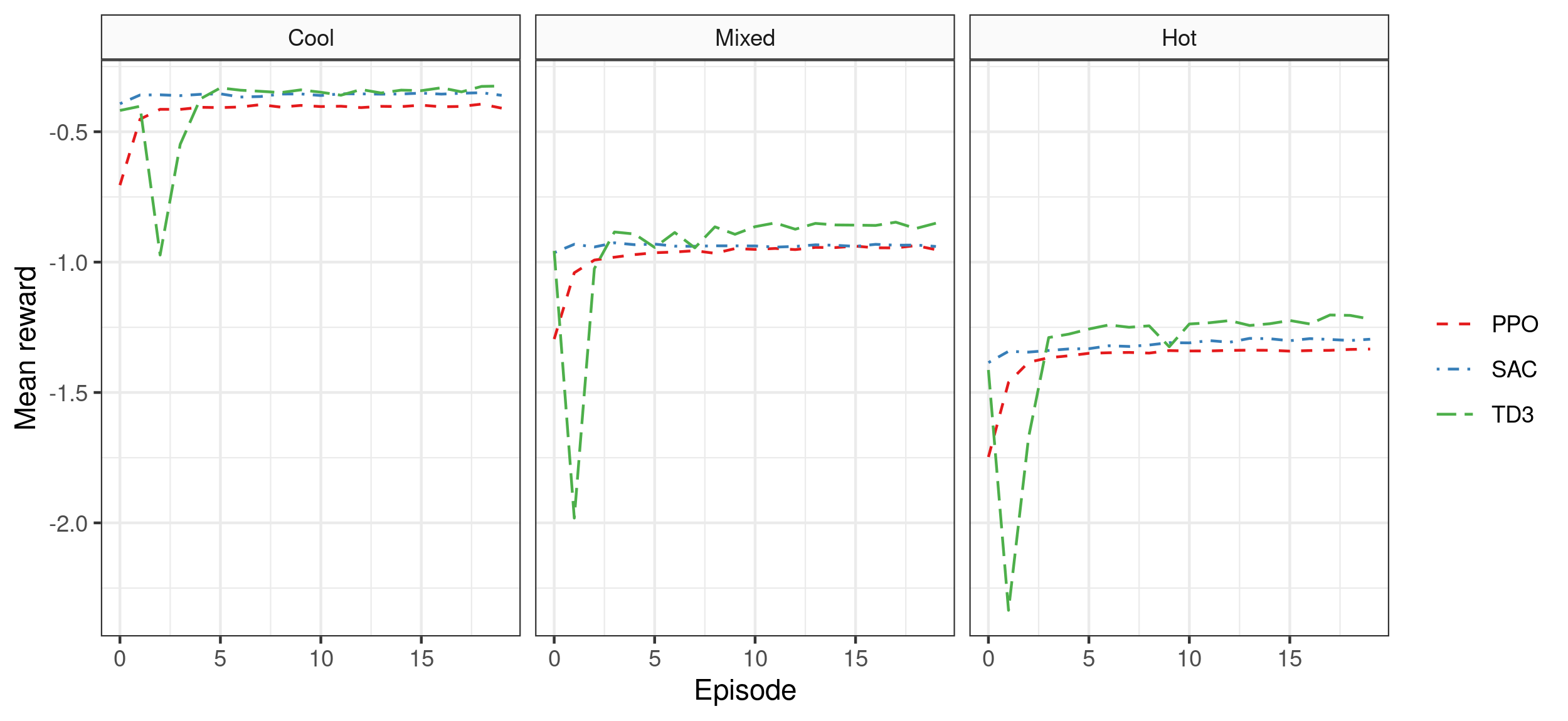}
    \caption{Mean episode rewards during 20 training episodes in \datacenter{}}
    \label{fig:train_rewards_datacenter}
\end{figure}

Regarding \datacenter{}, in Figure \ref{fig:train_rewards_datacenter} we can see a higher instability of TD3 in the first episodes, and a later better performance against SAC and PPO. In this environment, SAC again demonstrates early convergence, outperforming PPO in cool and hot climates, and equaling it in the mixed case.

\begin{figure}
    \centering
    \includegraphics[width=\linewidth]{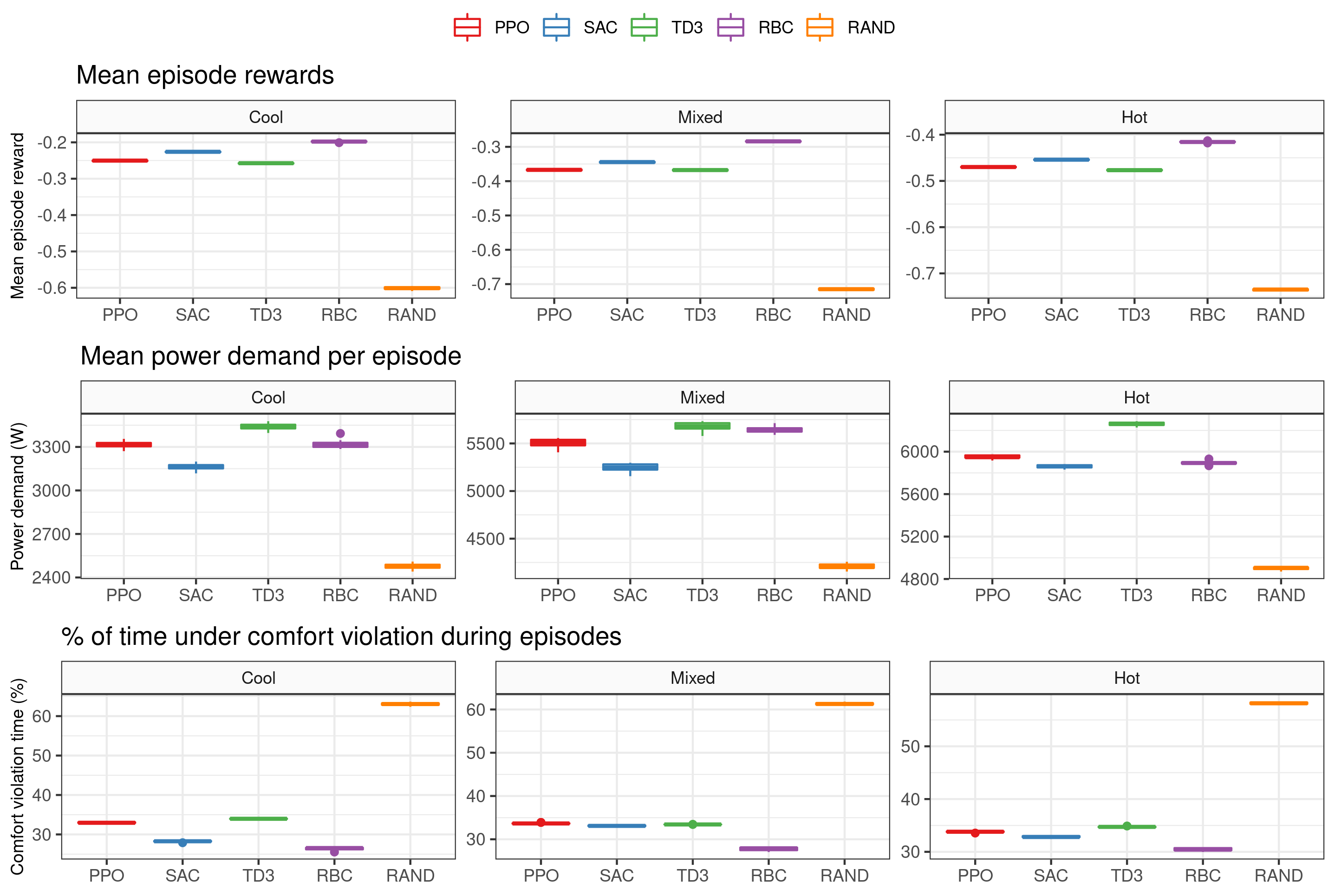}
    \caption{Results for 20 evaluation episodes in \fivezone{}}
    \label{fig:eval_metrics_5Zone}
\end{figure}

\begin{figure}
    \centering
    \includegraphics[width=\linewidth]{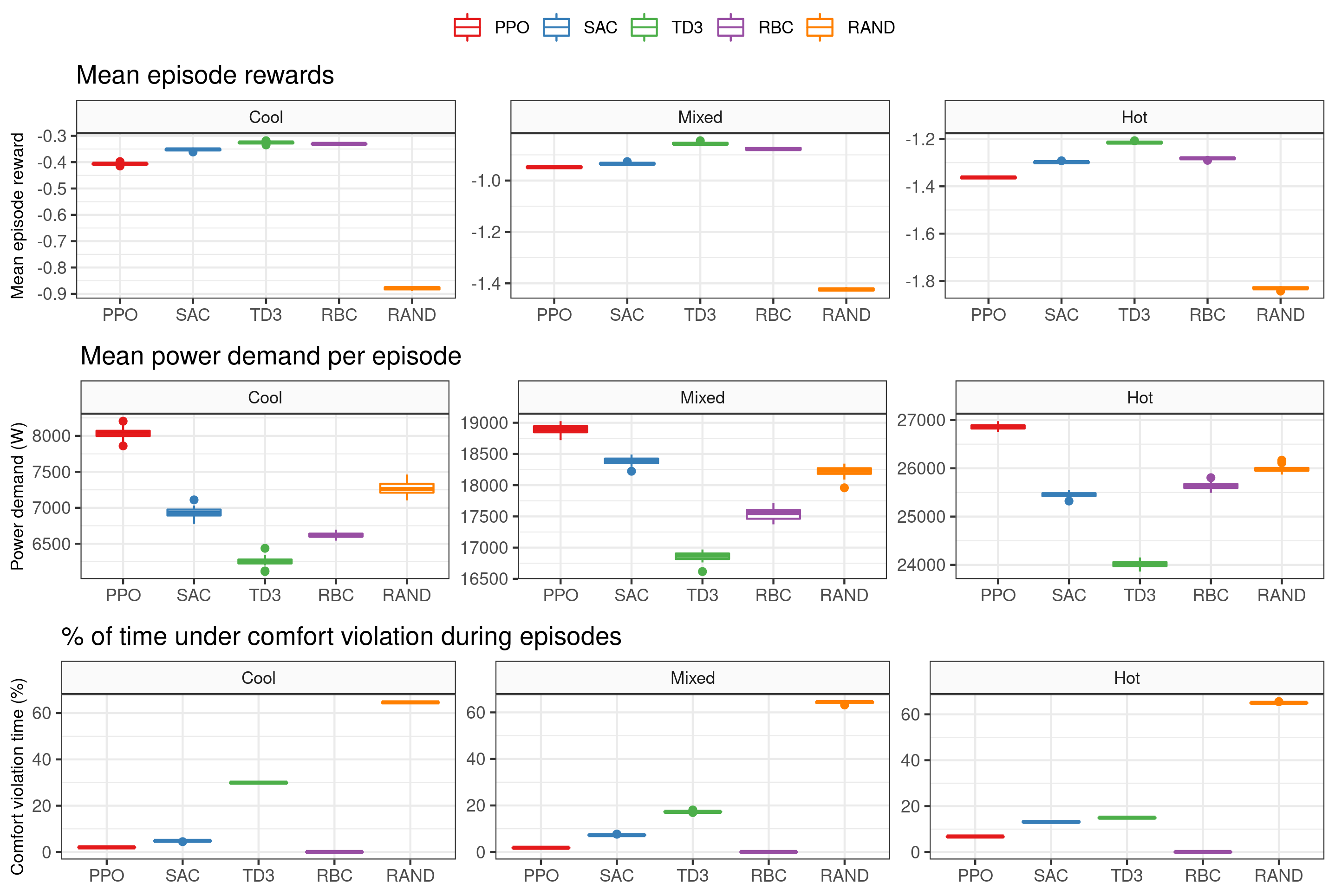}
    \caption{Results for 20 evaluation episodes in \datacenter{}}
    \label{fig:eval_metrics_datacenter}
\end{figure}

Once the algorithms were trained and reached convergence, they were evaluated during 20 episodes using the best model obtained during training. Figures \ref{fig:eval_metrics_5Zone} and \ref{fig:eval_metrics_datacenter} show the results obtained for each environment, considering the metrics already presented in section \ref{sec:5.1}: mean rewards per episode, power consumption and comfort violation. We will analyze the results obtained taking as baseline the performance of both RBCs and random agents (RAND). \edited{The values represented in the boxplots are summarized in Appendix \ref{sec:appendixB}.}

In the case of \fivezone{}, as shown in Figure \ref{fig:eval_metrics_5Zone}, the best comfort-consumption balance is guaranteed by the RBC. Being this an ad hoc solution implemented for each building, we are interested in knowing which DRL algorithm comes closest to its performance. In this case, as we could see during training, SAC is the closest agent to RBC control in all climates, followed by PPO and TD3, with similar mean rewards in all scenarios. Thus, none of the agents manages to outperform RBC regarding this metric.

Considering the comfort and consumption values, we observe that RBC's main competitor, SAC, manages to reduce power demand without major penalties in comfort violation, especially in hot weather. On the other hand, TD3 is the most energetically costly agent, while PPO maintains a significant balance between both metrics without demonstrating outstanding performance.

Looking at Figure \ref{fig:eval_metrics_datacenter}, corresponding to \datacenter{}, we find a higher competitiveness with respect to RBC, which is outperformed in all scenarios by TD3 in terms of reward. We see, in turn, significant power savings with respect to the other agents, albeit at the cost of a greater comfort violation. This could potentially be less interesting if we are dealing with a building where the temperature must always be kept in a specific range for safety reasons, so more conservative options such as SAC, PPO or RBC itself could be more suitable.

\begin{figure}
    \centering
    \includegraphics[width=\linewidth]{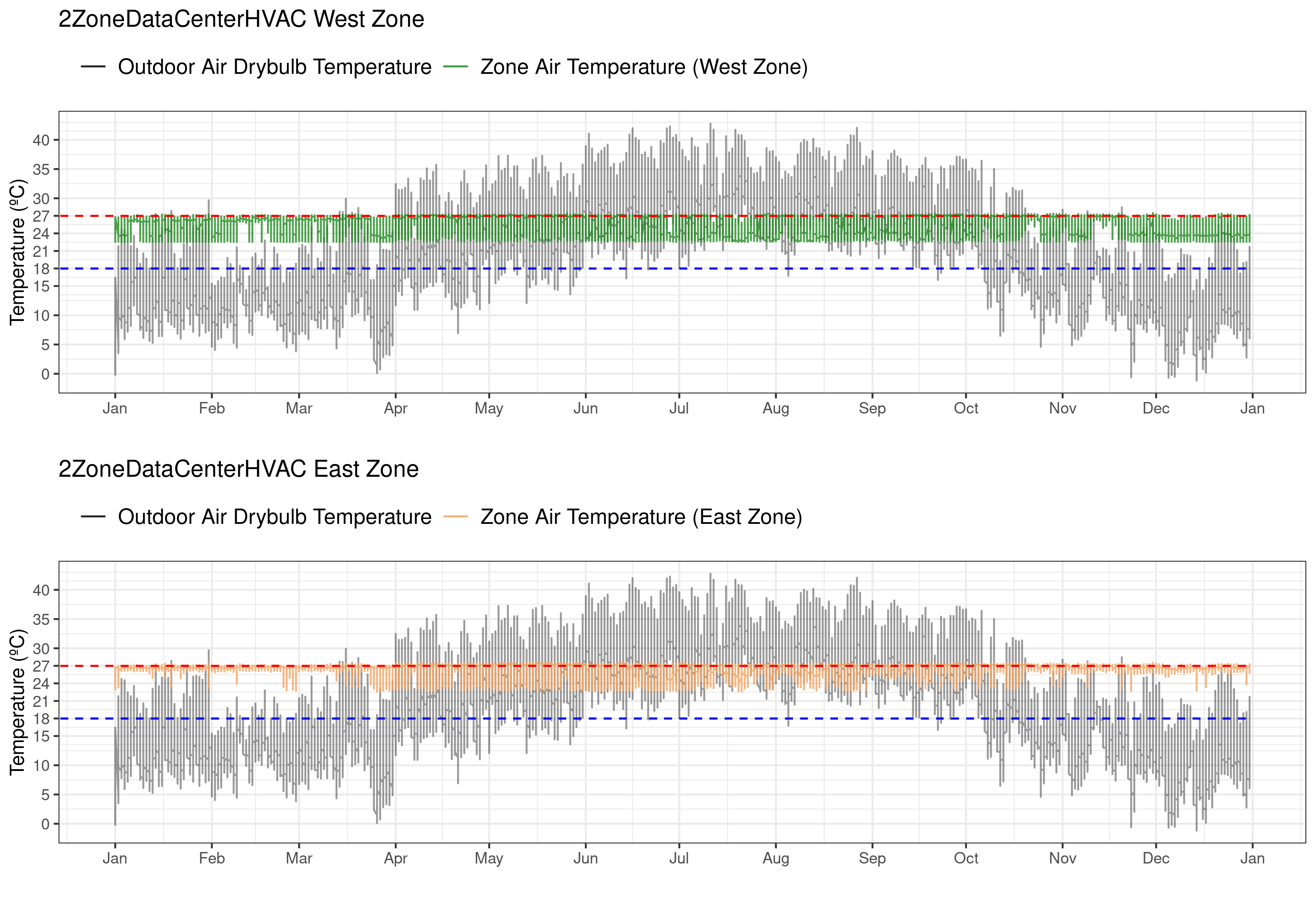}
    \caption{Evolution of temperatures during a year of simulation using TD3 in \datacenter{}-hot. Comfort thresholds are marked with the horizontal dotted red and blue lines}
    \label{fig:temperatures_TD3_datacenter_hot}
\end{figure}

\begin{figure}
    \centering
    \includegraphics[width=\linewidth]{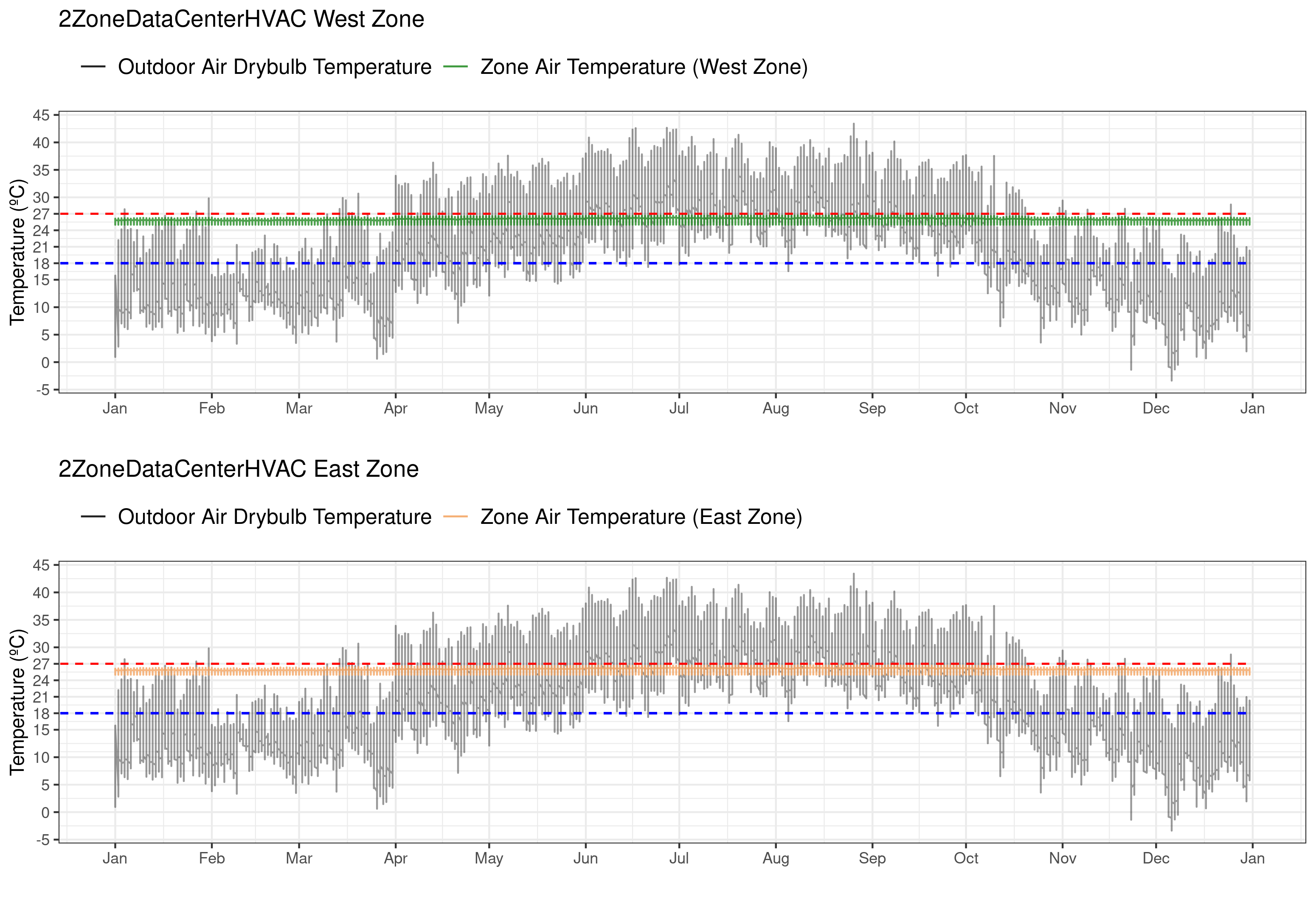}
    \caption{Evolution of temperatures during a year of simulation using RBC in \datacenter{}-hot. Comfort thresholds are marked with the horizontal dotted red and blue lines}
    \label{fig:temperatures_RBC_datacenter_hot}
\end{figure}

When we refer to conservative solutions, we mean those that offer indoor temperatures further away from the limits of comfort ranges. This makes it possible to deal with temperature inertia at the cost of investing more power in ensuring stable temperatures. For example, if we compare the indoor temperatures throughout the simulation for TD3 (Figure \ref{fig:temperatures_TD3_datacenter_hot}) and RBC (Figure \ref{fig:temperatures_RBC_datacenter_hot}), we observe that the former temperatures are closer to the upper comfort limit than those of the RBC. This behavior may be interpreted as the DRL agent trying to optimize consumption by taking a higher risk in ensuring comfort, thus leading to indoor temperatures closer to the limit allowed by the reward function.

A more feasible alternative in ensuring compliance with comfort limits might involve modifying the reward function used, giving greater importance to comfort versus consumption. This approach will be specifically addressed in section \ref{sec:5.5}.

\begin{figure}
    \centering
    \includegraphics[width=\linewidth]{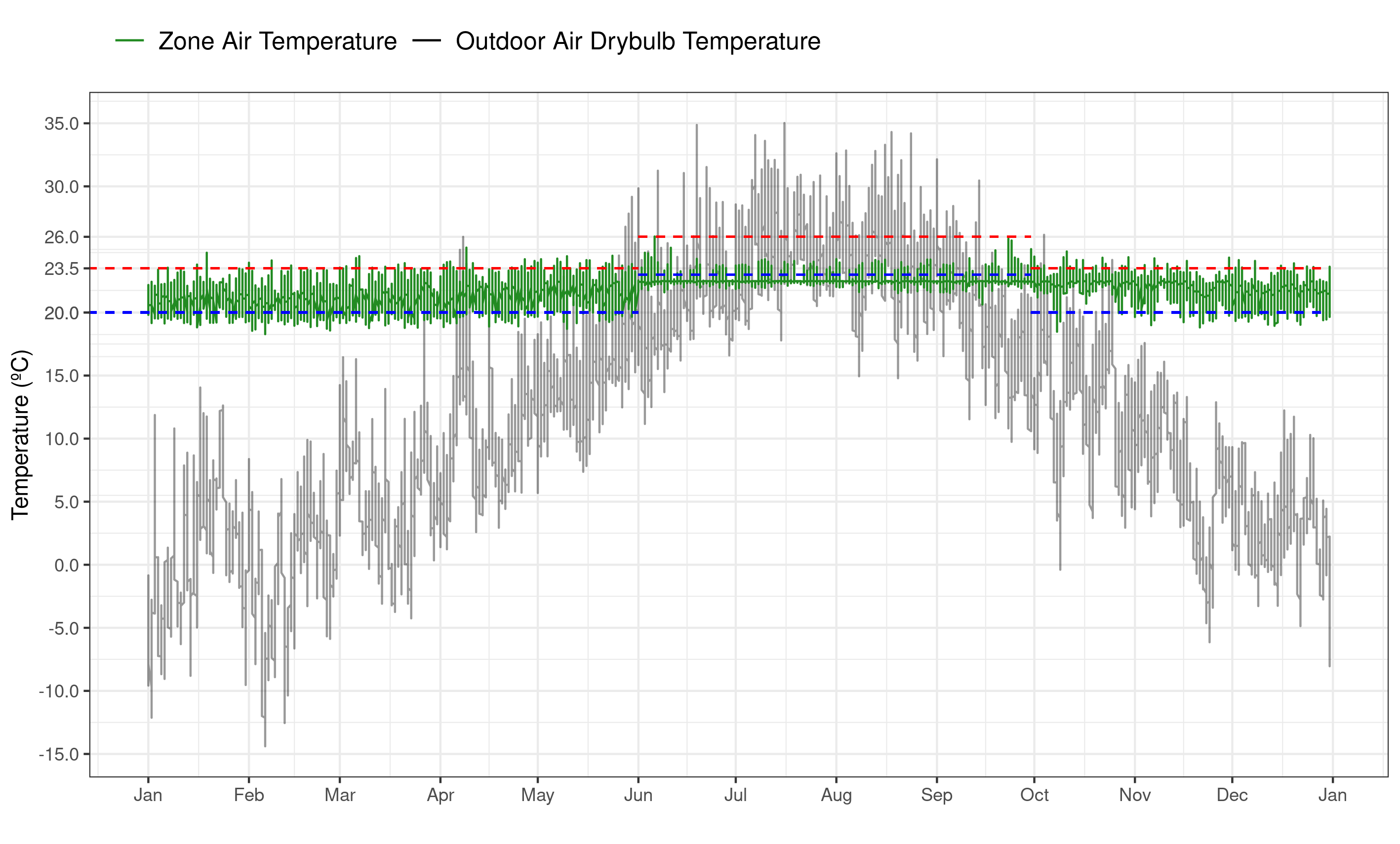}
    \caption{Evolution of temperatures during a year of simulation using SAC in \fivezone{}-mixed. Comfort thresholds are marked with the horizontal dotted red and blue lines}
    \label{fig:temperatures_SAC_5Zone_mixed}
\end{figure}

Finally, it is worth noting the difficulties of the DRL algorithms in adapting to comfort ranges that vary at specific periods of the episode. This is the case of \fivezone{} and the variations in the comfort ranges defined for the warm and cold months, as explained in section \ref{sec:5.1}. For instance, Figure \ref{fig:temperatures_SAC_5Zone_mixed} shows how the sudden change in the desired temperature range poses difficulties for SAC, which does not adapt well enough to the new comfort requirements.

Having provided a first insight into the performance of the different algorithms, in the following subsections we will address issues related to performance improvement and generalization capabilities of DRL agents.

\subsection{Robustness test}
\label{sec:5.3}

We will test the performance of agents executed in environments that differ from those in which they have been trained. This will enable to know how far an agent is able to generalize and extrapolate the knowledge acquired in a given climate to be applied in a different one. 

\begin{figure}
    \centering
    \includegraphics[width=\linewidth]{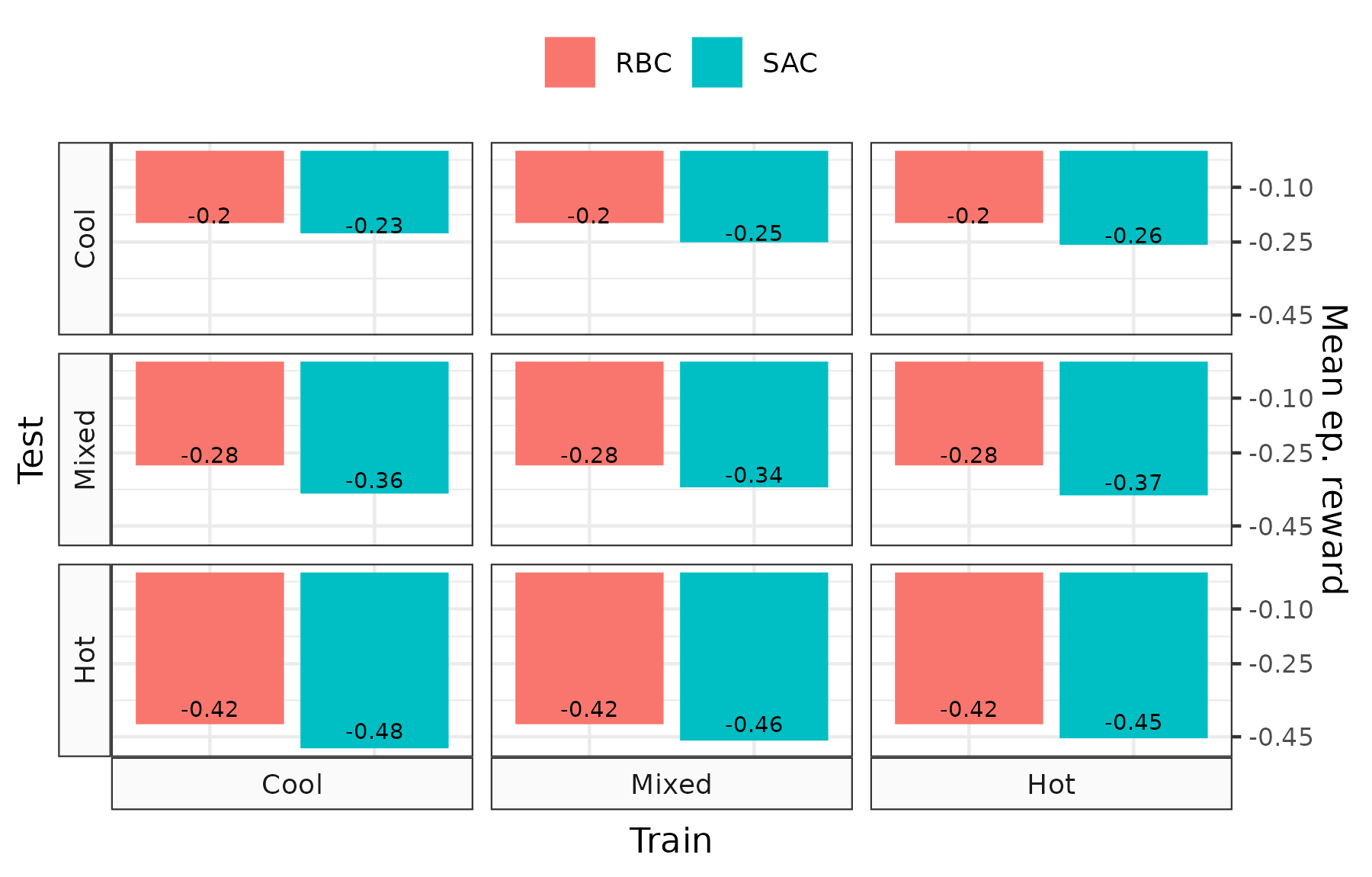}
    \caption{\edited{Evaluation rewards for SAC agents trained in different weathers. Compared with RBC mean rewards in \fivezone{}}}
    \label{fig:robustness_test}
\end{figure}

According to Figure \ref{fig:robustness_test}, we observe that the agents that perform best in each climate are those that have been trained in that same climate, as would be expected. \edited{After applying Mann-Whitney U tests, we obtain significant differences between the rewards obtained in each test environment ($p < 0.05$ in all cases).} 

In all cases, the following order of performance holds true: cool is the climate for which the greatest rewards are obtained, followed by mixed and hot. Even agents trained in mixed and hot get a higher reward in cool than in their respective training climates. The same happens for the agent trained in hot climate, whose reward is worst in its own climate than in the rest.

As a summary, Table \ref{tab:robustness-tests} gathers the results obtained for each SAC training and evaluation combination in \fivezone{}, which we will further discuss in section \ref{sec:6.2}.

\begin{table}[]
    \caption{\edited{Rewards for every training and evaluation combination using SAC in \fivezone{}}}
    \label{tab:robustness-tests}
    \footnotesize
    \centering
    \begin{tabular}{ccccc}
    \cline{3-5}
                  &                & \multicolumn{3}{c}{\textbf{Train}}                                                                                                                                                                                 \\ \cline{3-5} 
                  &                & \textbf{Cool}                                                        & \textbf{Mixed}                                                       & \textbf{Hot}                                                         \\ \hline
    \textbf{}     & \textbf{Cool}  & \begin{tabular}[c]{@{}c@{}}\textbf{mean} = -0.226\\ \textbf{sd} = 0.00129\end{tabular} & \begin{tabular}[c]{@{}c@{}}\textbf{mean}= -0.251\\ \textbf{sd} = 0.00129\end{tabular}  & \begin{tabular}[c]{@{}c@{}}\textbf{mean} = -0.258\\ \textbf{sd} = 0.00135\end{tabular} \\ \cline{2-5} 
    \textbf{Test} & \textbf{Mixed} & \begin{tabular}[c]{@{}c@{}}\textbf{mean} = -0.361\\ \textbf{sd} = 0.00217\end{tabular} & \begin{tabular}[c]{@{}c@{}}\textbf{mean} = -0.344\\ \textbf{sd} = 0.00191\end{tabular} & \begin{tabular}[c]{@{}c@{}}\textbf{mean} = -0.366\\ \textbf{sd} = 0.00198\end{tabular} \\ \cline{2-5} 
    \textbf{}     & \textbf{Hot}   & \begin{tabular}[c]{@{}c@{}}\textbf{mean} = -0.482\\ \textbf{sd} = 0.0012\end{tabular}  & \begin{tabular}[c]{@{}c@{}}\textbf{mean} = -0.46\\ \textbf{sd} = 0.00117\end{tabular}  & \begin{tabular}[c]{@{}c@{}}\textbf{mean} = -0.454\\ \textbf{sd} = 0.00111\end{tabular} \\ \hline
    \end{tabular}
\end{table}

\subsection{Sequential learning}
\label{sec:5.4}

Another objective proposed in this work is to address the application of sequential learning: an approach that involves the progressive training of an agent under different weather conditions. On this basis, our goal in applying this incremental learning is to verify whether an agent trained progressively through different environments is capable of achieving better, or at least similar, performance to an agent trained in a single problem. 


Thus, based on the results obtained in section \ref{sec:5.2}, we will test to compare the performance of an agent directly trained on a single climate with one trained on every available climate. We will use SAC under the same configurations described in the previous experiments. The sequence of training considered is: (1) cool, (2) mixed, and (3) hot. 

\begin{figure}
    \centering
    \includegraphics[width=\linewidth]{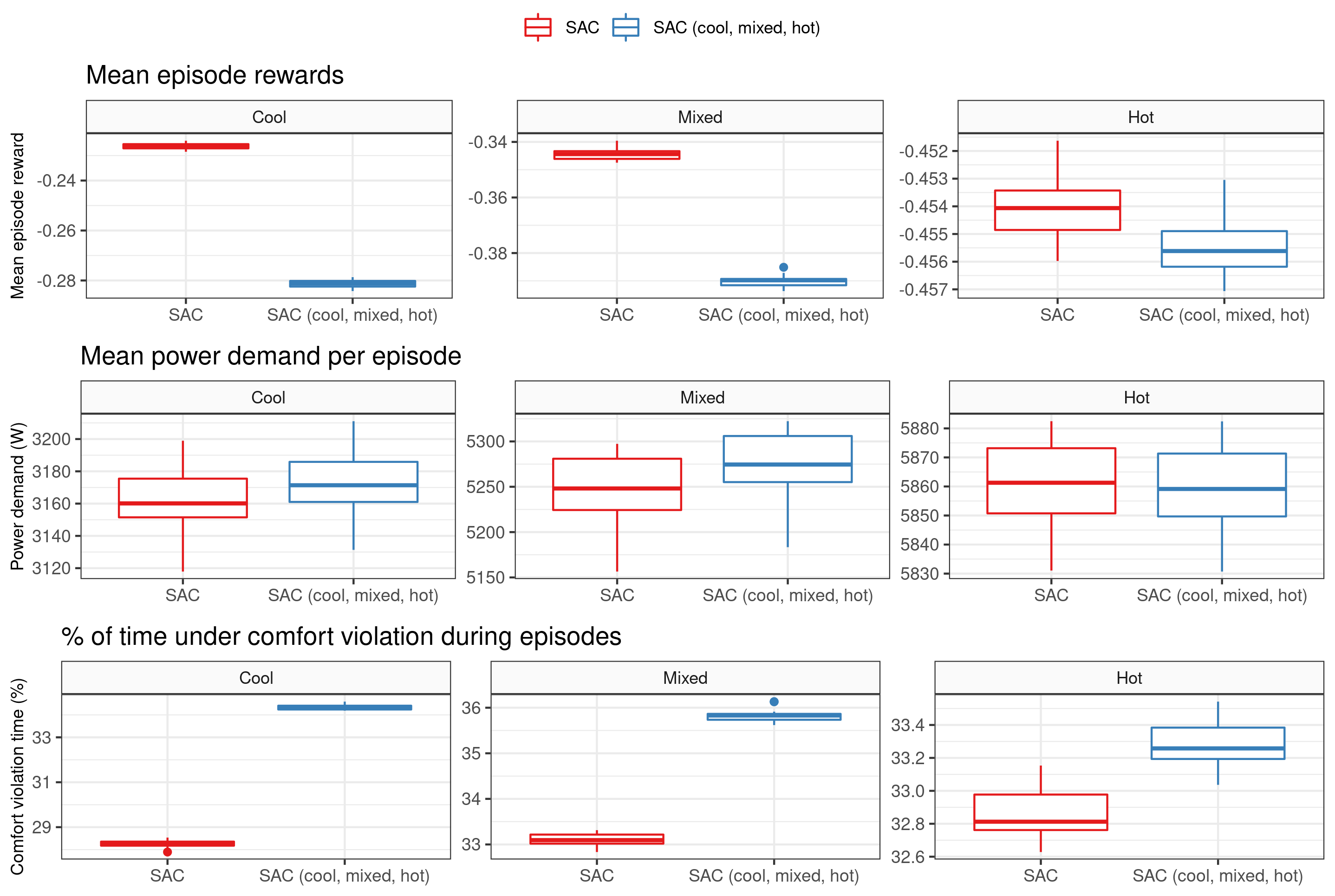}
    \caption{Results for SAC and sequential learning SAC in \fivezone{}}
    \label{fig:seq_learning}
\end{figure}

Therefore, in Figure \ref{fig:seq_learning} we observe that the performance of SAC trained under sequential learning is mostly inferior to that of standard SAC according to the metrics proposed.

It is noteworthy that the best performance obtained by SAC with sequential learning occurs in hot weather, which is actually the last weather used to train the agent. This could lead us to believe that the phenomenon of ``catastrophic forgetting'' \cite{french1999} might occur, as we will discuss in section \ref{sec:6.3}.

\subsection{Comfort-consumption trade-off}
\label{sec:5.5}

As we already anticipated in the experimentation of section \ref{sec:5.2}, the reward function chosen, as well as the weights given to comfort and consumption in the training of the agents should influence their performance.

In this section, we will compare how different weights for comfort and consumption influence the performance of the agents under the same rest conditions. Specifically, building on the results described in section \ref{sec:5.2}, we will use the best performing agents on each building ---SAC in \fivezone{}, and TD3 in \datacenter{}--- to compare the performance of the following agents:

\begin{itemize}
    \item An agent trained only on the basis of the comfort requirements ($\omega = 1$ in Equation \ref{eq:linear-reward}), regardless of consumption.
    \item Another trained with a weight of 75\% for consumption and 25\% for comfort ($\omega = 0.25$).
    \item An agent trained with the same weight for consumption and comfort ($\omega = 0.5$).
    \item And, finally, an agent trained with a weight of 75\% for comfort and 25\% for consumption ($\omega = 0.75$).
\end{itemize}

\begin{figure}
    \centering
    \includegraphics[width=\linewidth]{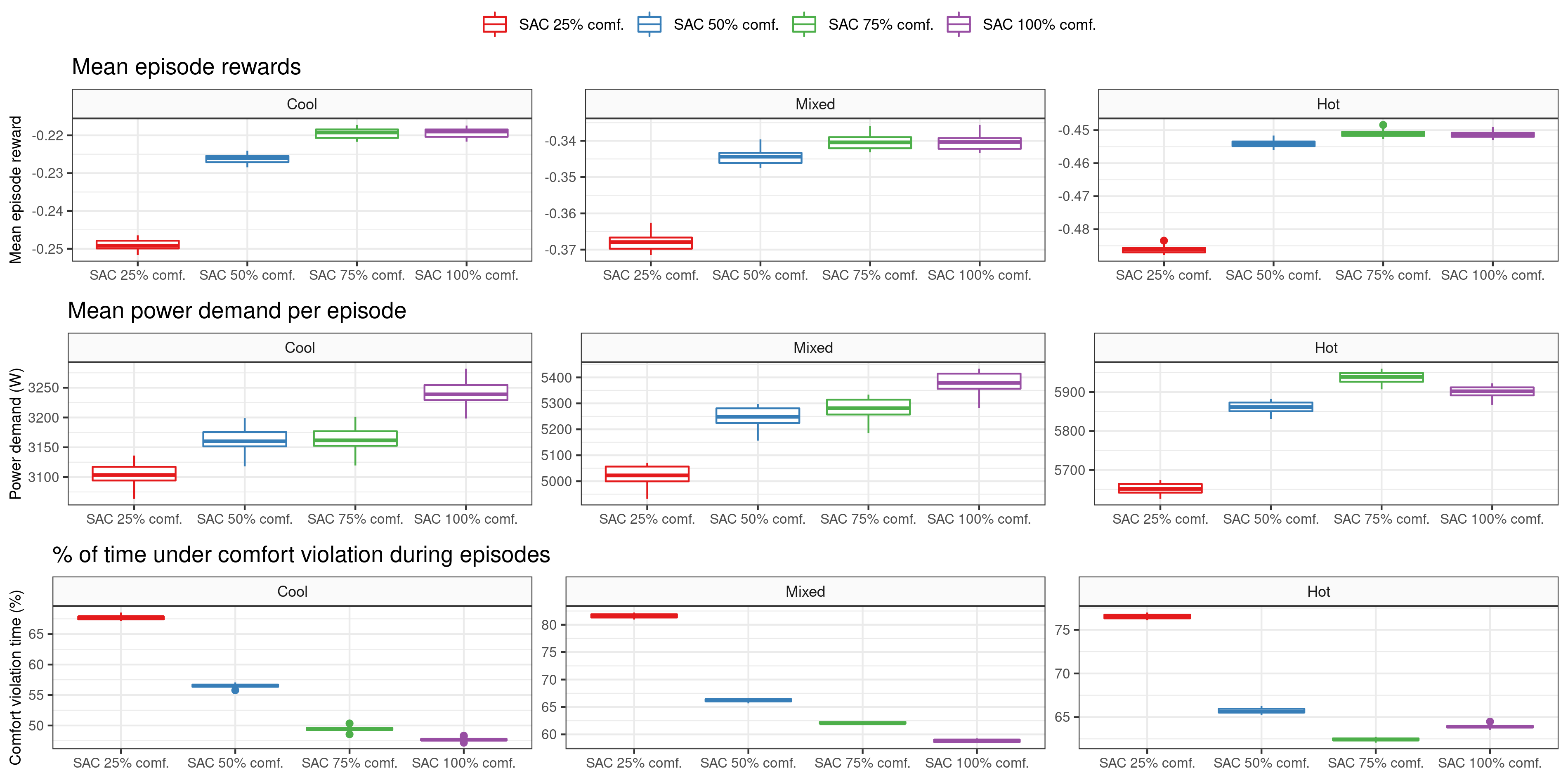}
    \caption{Results for SAC with different comfort weights in \fivezone{}}
    \label{fig:sac_tradeoff_metrics}
\end{figure}

\begin{figure}
    \centering
    \includegraphics[width=\linewidth]{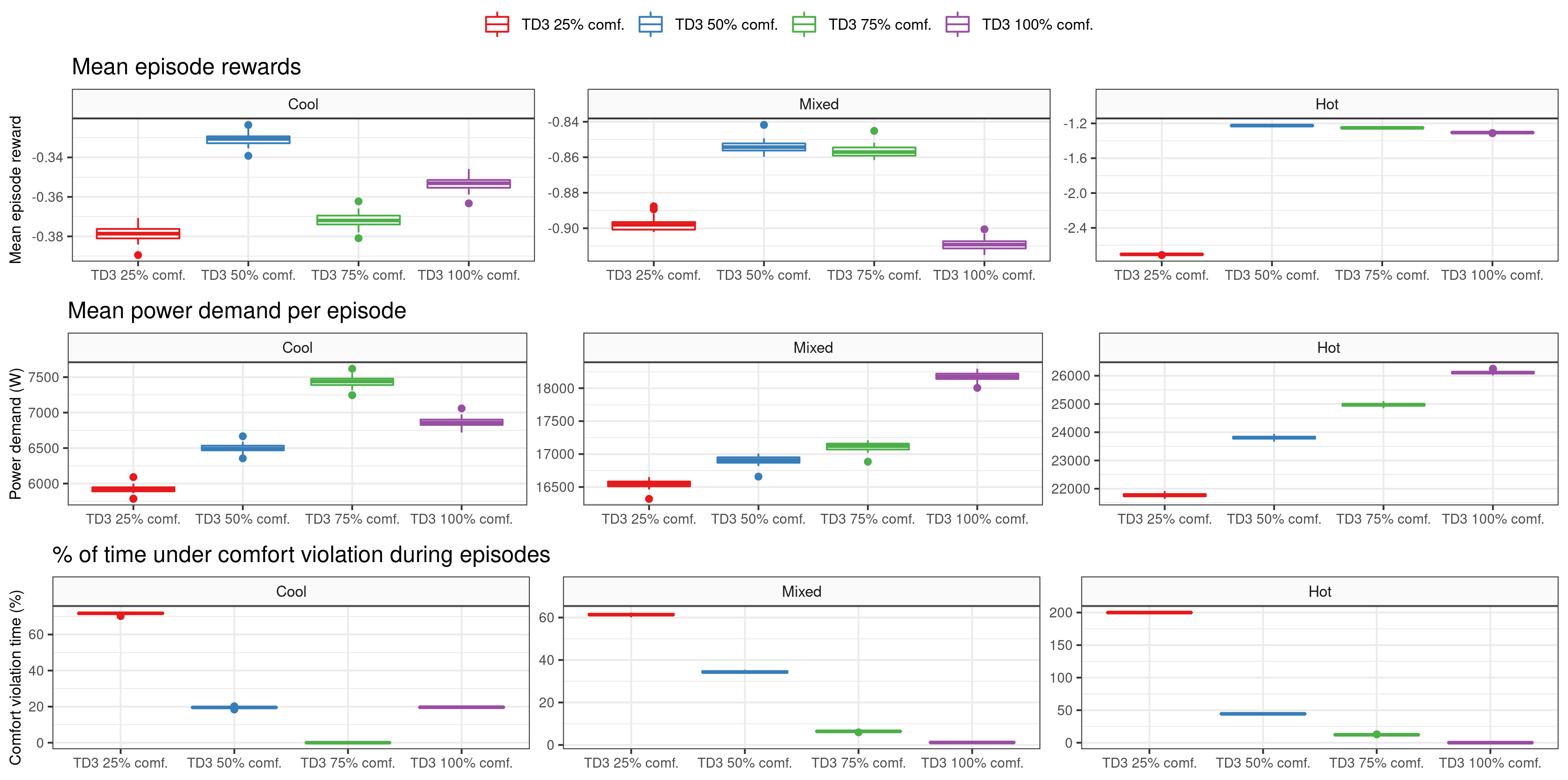}
    \caption{Results for TD3 with different comfort weights in \datacenter{}}
    \label{fig:td3_tradeoff_metrics}
\end{figure}

Thus, the results obtained are shown in Figures \ref{fig:sac_tradeoff_metrics} and \ref{fig:td3_tradeoff_metrics}. From the figures we observe that a greater emphasis on comfort in the reward function reduces comfort violations, as would be expected. Improvements are especially significant from 25\% to 50\% comfort weight, and decrease as the comfort weight increases (although always improving).
    
However, there are two exceptions where assigning 100\% weight to comfort means some worsening from 75\%. This is the case for SAC in hot climate and TD3 in cool climate. In section \ref{sec:6.4} we will discuss possible causes for these exceptions.
    
Finally, improving comfort means ---apart from the exceptions mentioned above--- an increase in power demand. In this multi-objective problem, the trade-off is disrupted as soon as one objective is given greater importance than the other.

\section{Discussion}
\label{sec:6}

In this section, we will discuss the results obtained after experimentation, not only describing them but also identifying gaps to be addressed in future work.

\subsection{Rewards and comparison between algorithms}
\label{sec:6.1}

We begin by analyzing the results obtained by the different control algorithms in the set of proposed environments. \edited{Overall, the results obtained are consistently stable with minimal variation over the 20 evaluation episodes. Stable results from multiple evaluation episodes allows us to confirm that there are no random factors significantly affecting the agents' results (see Appendix \ref{sec:appendixB})}.

As shown in section \ref{sec:5.1}, in the case of \fivezone{}, no DRL algorithm was able to outperform RBC considering a reward which gives equal importance to comfort and consumption (50\% -- 50\%). It should also be noted that the buildings used in this experiment have a relatively simple control, which can be easily translated into a limited set of rules. However, as we discussed in the Experiments section, the results are particularly promising when we consider that there has been neither an exhaustive selection of observed variables nor an in-depth choice of hyperparameters.

In Figure \ref{fig:temperatures_SAC_5Zone_mixed} we also observe how DRL algorithms have problems in adapting to changing setpoints in certain periods of the simulation, such as the warm months. If we also consider that the higher the temperature, the more difficult it is to guarantee the comfort-consumption balance (see rewards for hot weather environments in Figure \ref{fig:eval_metrics_5Zone}), the loss of performance is self-evident. This loss of performance can be caused by external factors that affect the agent's ability to act, such building characteristics or an increased power demand from the HVAC system for cooling.

On the contrary, looking at the results in \datacenter{}, TD3 manages to outperform RBC in terms of average reward. The fact that a DRL agent is able to outperform an RBC in \datacenter{}, but not in \fivezone{} could be justified by the greater complexity of the control problem, as the latter may require a more sophisticated strategy, involving a larger number of control variables.

As we anticipated earlier, the main reason why TD3 gets a better reward is because of the large reduction in consumption compared to RBC. However, this seems to imply a higher comfort violation, which is ultimately profitable for the agent.

\begin{figure}
    \centering
        \begin{subfigure}{0.6\textwidth}
            \includegraphics[width=\textwidth]{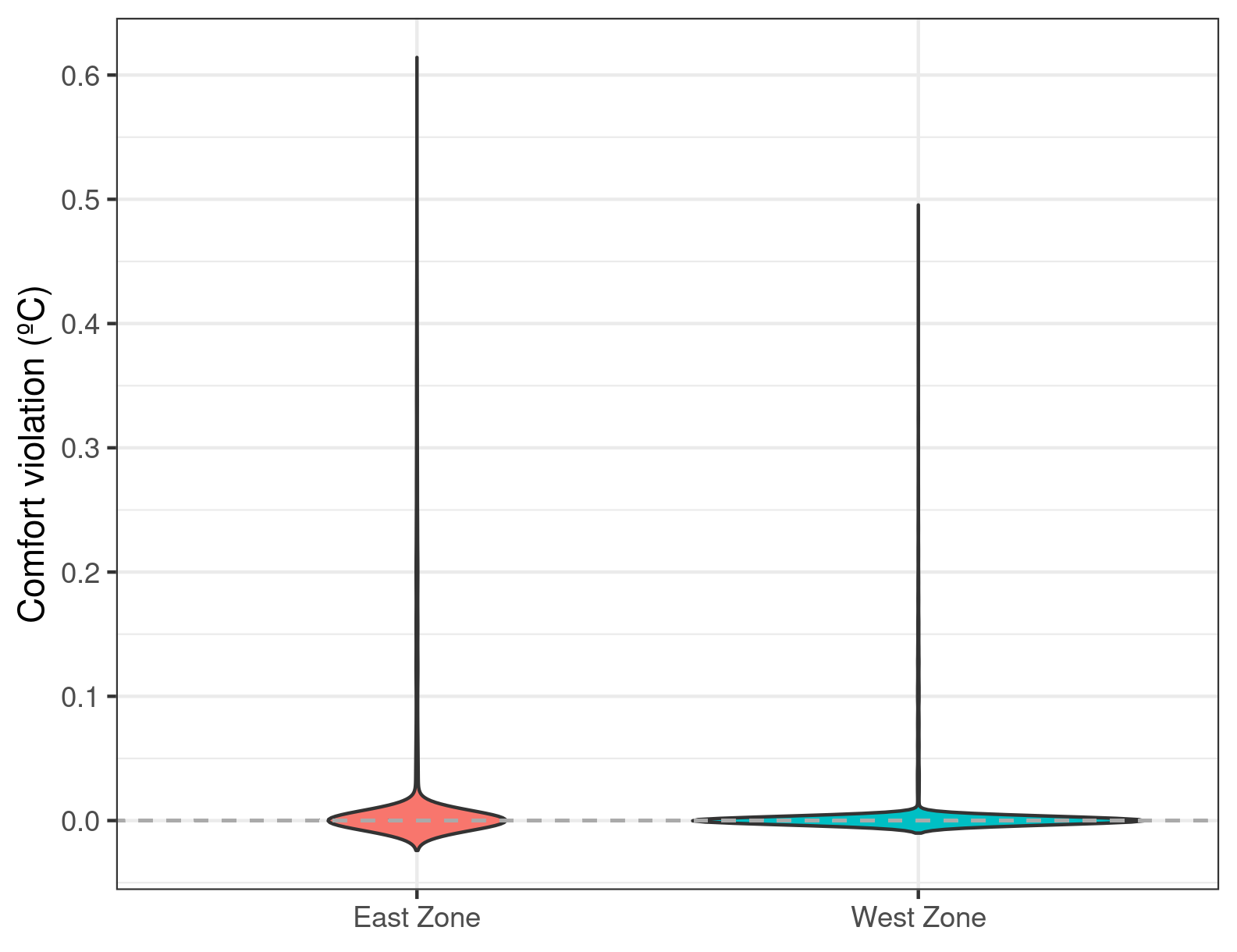}
            \caption{TD3 temperature deviations in \datacenter{} (hot weather)}
            \label{fig:comfort_violations_TD3_datacenter_hot}
        \end{subfigure}
        \begin{subfigure}{0.6\textwidth}
            \includegraphics[width=\textwidth]{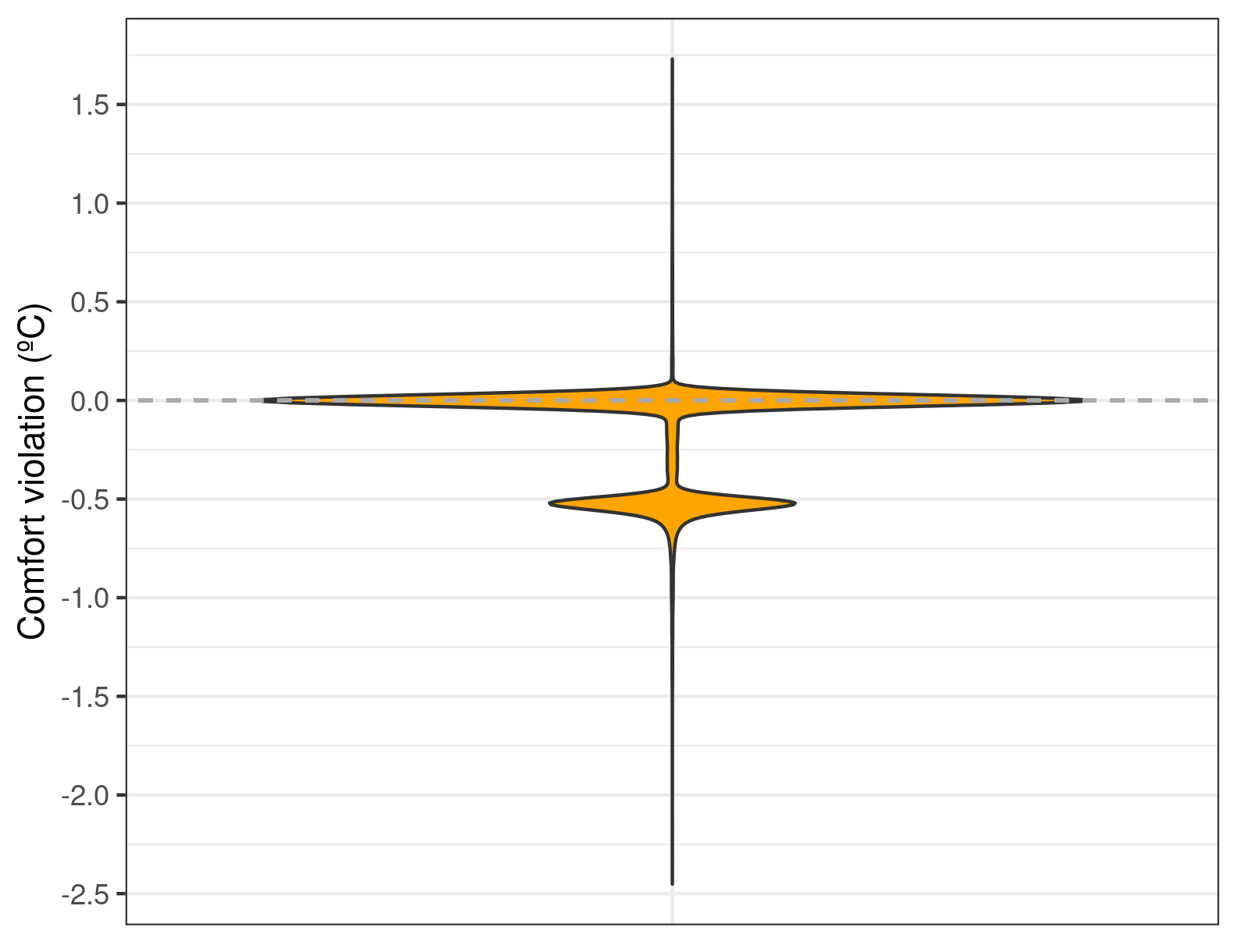}
            \caption{SAC temperature deviations in \fivezone{} (mixed weather)}
            \label{fig:comfort_violations_SAC_5Zone_mixed}
        \end{subfigure}   
    \caption{Comfort violations by TD3 and SAC in different scenarios}
    \label{fig:comfort_violations}
\end{figure}

An important aspect to question at this point is under what conditions we consider a comfort violation to occur. If we look at Figure \ref{fig:temperatures_TD3_datacenter_hot}, we can see that although many comfort violations occur for TD3 throughout the year, their value is quite low, mainly due to temperature inertia combined with the fact that power savings imply approaching comfort limits. Therefore, if we only look at Figure \ref{fig:eval_metrics_datacenter}, agnosticism towards real temperatures and deviations could lead us to believe that TD3 significantly sacrifices comfort in exchange for reducing consumption, thus reaching critical temperatures, which is not the case. As a proof, Figure \ref{fig:comfort_violations_TD3_datacenter_hot} shows the temperature violations achieved by TD3 with respect to the comfort ranges in \datacenter{}, proving that the temperature deviations exceed the upper comfort limit by less than 0.7\textcelsius, and never the lower limit.

Something similar occurs in the case of the SAC in \fivezone{}. Looking at Figure \ref{fig:comfort_violations_SAC_5Zone_mixed}, we observe that the comfort violations above the upper limit reach a maximum of about 1.6\textcelsius, while the comfort violations below the lower limit reach about 2.4\textcelsius. In the latter case, the comfort violations are higher because the agent adapts with difficulty to the change of the comfort range during summer months, usually staying below the desired minimum temperature.


Finally, we should not overlook a common factor related to the reward function used for both buildings: the value of the scaling factors $\lambda_P$ and $\lambda_T$ (see Equation \ref{eq:reward}), which were chosen empirically. This may lead to slight inaccuracies in the reward calculation. Nevertheless, assuming the same values for all algorithms allows comparing the results under similar conditions.

\subsection{Robustness and generalization}
\label{sec:6.2}

The results of the robustness and generalisation experiments yield some interesting insights. On the one hand, we confirm that the agents that performed best in a given environment were those that had been trained in the same environment. That is to say, an agent executed in an environment different from the one in which it was trained reported a loss of performance in terms of average reward.

On the other hand, the rewards obtained in each test environment (see Figure \ref{fig:robustness_test}) show that the climate associated with the highest reward was cool, followed by mixed and hot. However, these values should be interpreted with caution: the fact that the rewards obtained in the hot climate were lower for each agent than in the cool climate does not necessarily mean that the agents are unable to perform properly in hot environments, but may be due to external factors. For example, a reward of around -0.4 in hot weather could be comparable to one of around -0.2 in cool weather simply because the HVAC system consumes less when heating than when cooling, which is beyond the agents' ability to act. This leads us to suggest that a column-by-column study of Figure \ref{fig:robustness_test} may not be as informative as a row-by-row comparison of the rewards.

These results also lead us to question whether a specific and fixed climate is the best approach when training DRL algorithms in this domain. Future work should try to clarify whether the differences in the control exerted by agents trained in different climates are really significant ---we are currently dealing with centesimal variations in the value of the mean rewards---, as well as to address alternatives to training based on single climates. For example, an alternative may involve using a dataset which pools records from several different climates, and from which one climate observation is randomly sampled at each time step. This is an idea already addressed in other works, such as \cite{du2021}, where a DRL agent was trained by sampling observations of heating and cooling scenarios, thus leading to more adaptive action strategies. Consequently, this solution may greatly enhance agents' generalization during training.

\subsection{Effectiveness of sequential learning}
\label{sec:6.3}

The search for an improvement in training by applying sequential learning also raises important issues in its application and results. As shown in Figure \ref{fig:seq_learning}, this approach resulted in a performance degradation with respect to standard SAC in each of the climates. Thus, the proposed method did not lead to improvements in expected performance.

A noticeable finding is that the results obtained report a lower loss of performance by the agent trained with sequential learning in the hot climate. This leads us to question whether this is caused by the fact that it was the last climate used for training. If so, we would be facing the already mentioned problem of catastrophic forgetting, which implies a bias in the agent's learning towards the most recent training data, discarding the learning of previous climates.

\begin{figure}
    \centering
    \includegraphics[width=\linewidth]{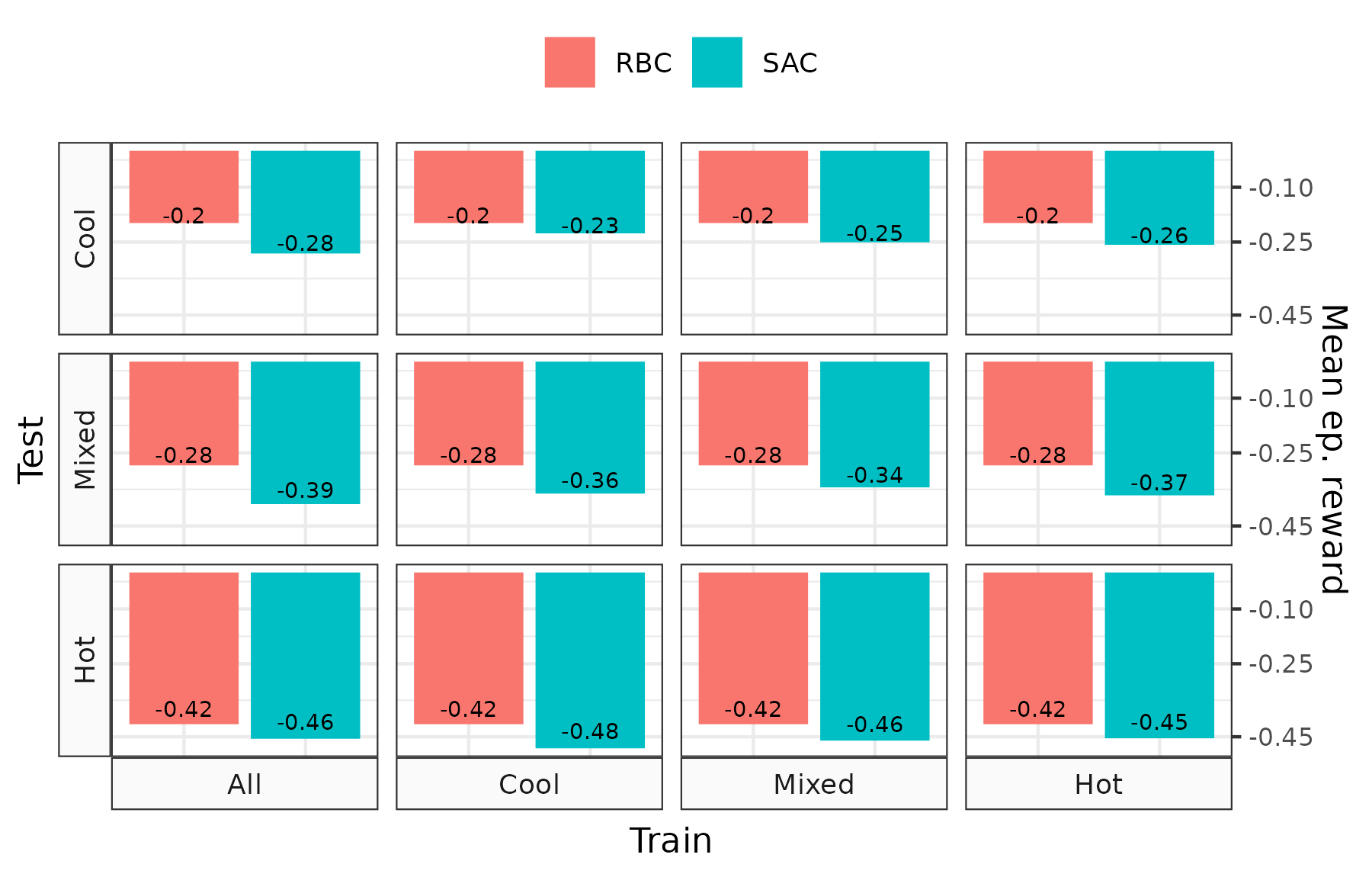}
    \caption{\edited{Evaluation rewards for SAC trained with sequential learning in all climates, and SAC trained in single climates. Compared with RBC mean rewards in \fivezone{}}}
    \label{fig:climates_SAC_comparison}
\end{figure}

In fact, if we compare SAC trained with sequential learning with the rest of SAC agents trained in single climates (Figure \ref{fig:climates_SAC_comparison}), we notice that the rewards are always worse, with the exception of hot, where results are close to those of SAC trained in mixed, and superior to those trained in cool. Again, this improved control in hot climate could be a direct consequence of catastrophic forgetting, as it concerns the last climate used in the agent's training, thus partially forgetting what the agent learned for cool and mixed environments.

\begin{figure}
    \centering
    \includegraphics[width=\linewidth]{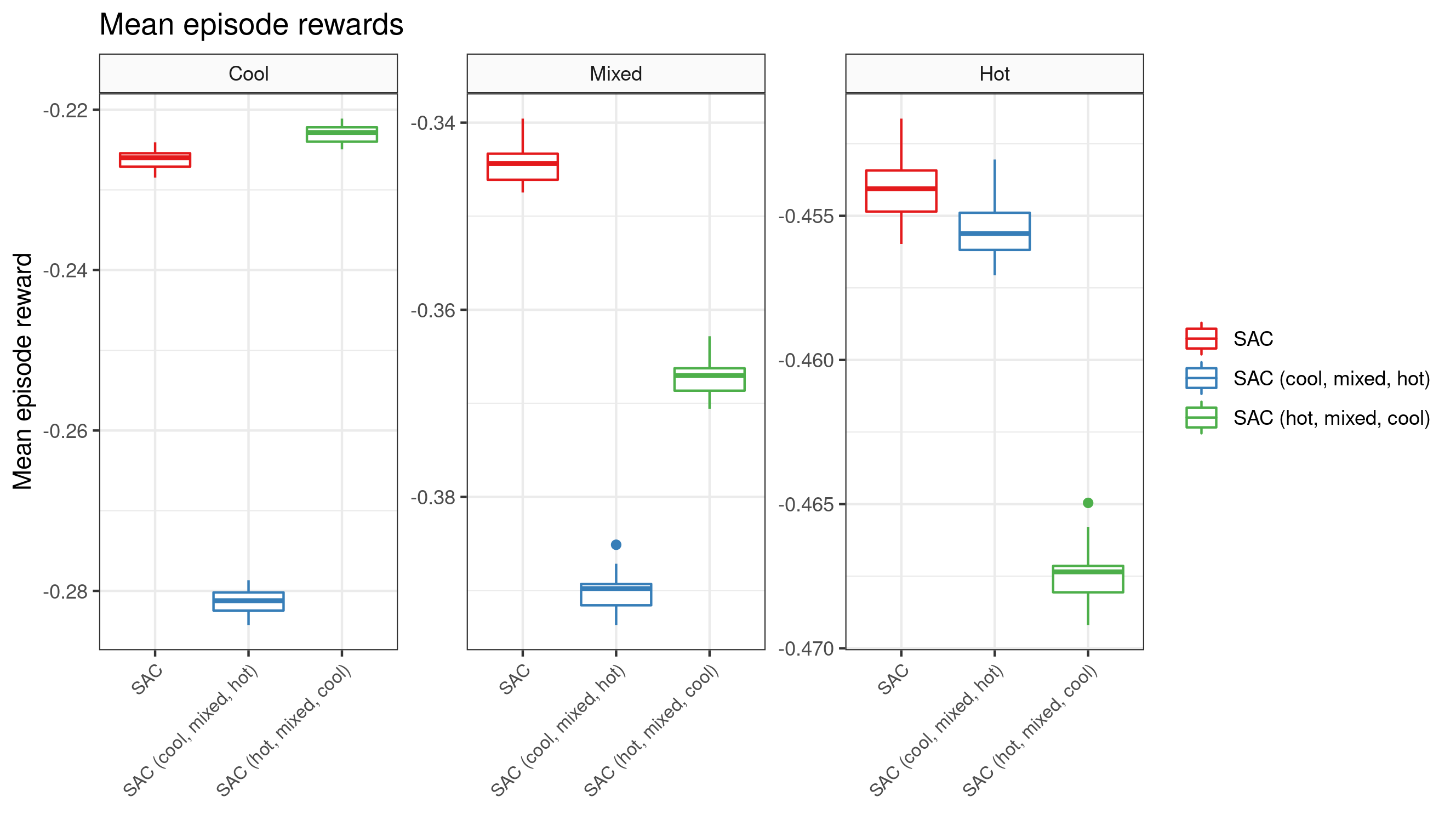}
    \caption{Comparison between standard SAC rewards, and SAC agents trained under sequential learning in different orders}
    \label{fig:cv_learning_2}
\end{figure}

To evaluate whether this phenomenon occurs, we tested the performance of an agent trained with sequential learning by following the reverse process. To do so, a SAC agent was firstly trained in hot weather, then in mixed, and finally in cool. Its performance is shown and compared with the rest of the SAC agents in Figure \ref{fig:cv_learning_2}, where we can observe a significant improvement in the reward for cool weather, at the expense of worsening performance for the rest of climates.

Looking at the results reinforces the idea that there is a bias towards the latter training climate --- in this case, cool. This leads us to believe that catastrophic forgetting does indeed occur, preventing diversified training.

It is also noteworthy how the new agent trained in all three climates manages to outperform a SAC agent trained only in cold, so the application of sequential learning may not be worthwhile in these circumstances. However, as a consequence of the existence of catastrophic forgetting, future work should address the influence of the architecture of the neural networks used by the agents, as well as the way in which the training data is handled. Again, we reiterate the idea that sampling from a set of observations from different climates may be a suitable option for this type of problems.

\subsection{Comfort-consumption trade-off}
\label{sec:6.4}

From the results shown in Figures \ref{fig:sac_tradeoff_metrics} and \ref{fig:td3_tradeoff_metrics}, we can see how the weight assigned to comfort ---and, therefore, to consumption--- directly influences the importance of this value in the training and final performance of the algorithms.

As we already anticipated, a higher comfort assurance implies, in most cases, a consequent increase in power demand. However, there are interesting cases such as SAC in \fivezone{} and cold weather, where it is possible to reduce the comfort violation by increasing its importance in the reward function without significant drawbacks in terms of consumption.

Furthermore, the choice of weighting for comfort and consumption depends entirely on the problem and the environment we are dealing with. However, there are cases, such as the one mentioned above, where we can achieve significant improvements in comfort with hardly any penalty in consumption. It is therefore a problem that needs to be approached cautiously and, if possible, several options should be evaluated in a simulated way.

If we look at the rewards obtained by the different SAC agents in \fivezone{}, we observe that a greater emphasis on comfort generally leads to better overall performance. As we have observed, this is because improvements in comfort do not have a large impact on power consumption, which offsets the results.

Nevertheless, in the case of TD3 in \datacenter{}, the comfort and consumption weighting changes do pose more of a challenge in terms of trade-off, with 50\%-50\% being the best weighting choice in all climates.

Let us now study what happens with the two exceptions mentioned in section \ref{sec:5.5}: SAC in \fivezone{} and hot weather, and TD3 in \datacenter{} with cool weather, where a 75\% weight for comfort in the reward function gives better results in terms of comfort violations than assigning a 100\% percent weight to this value.

\begin{figure}
    \centering
    \includegraphics[width=\linewidth]{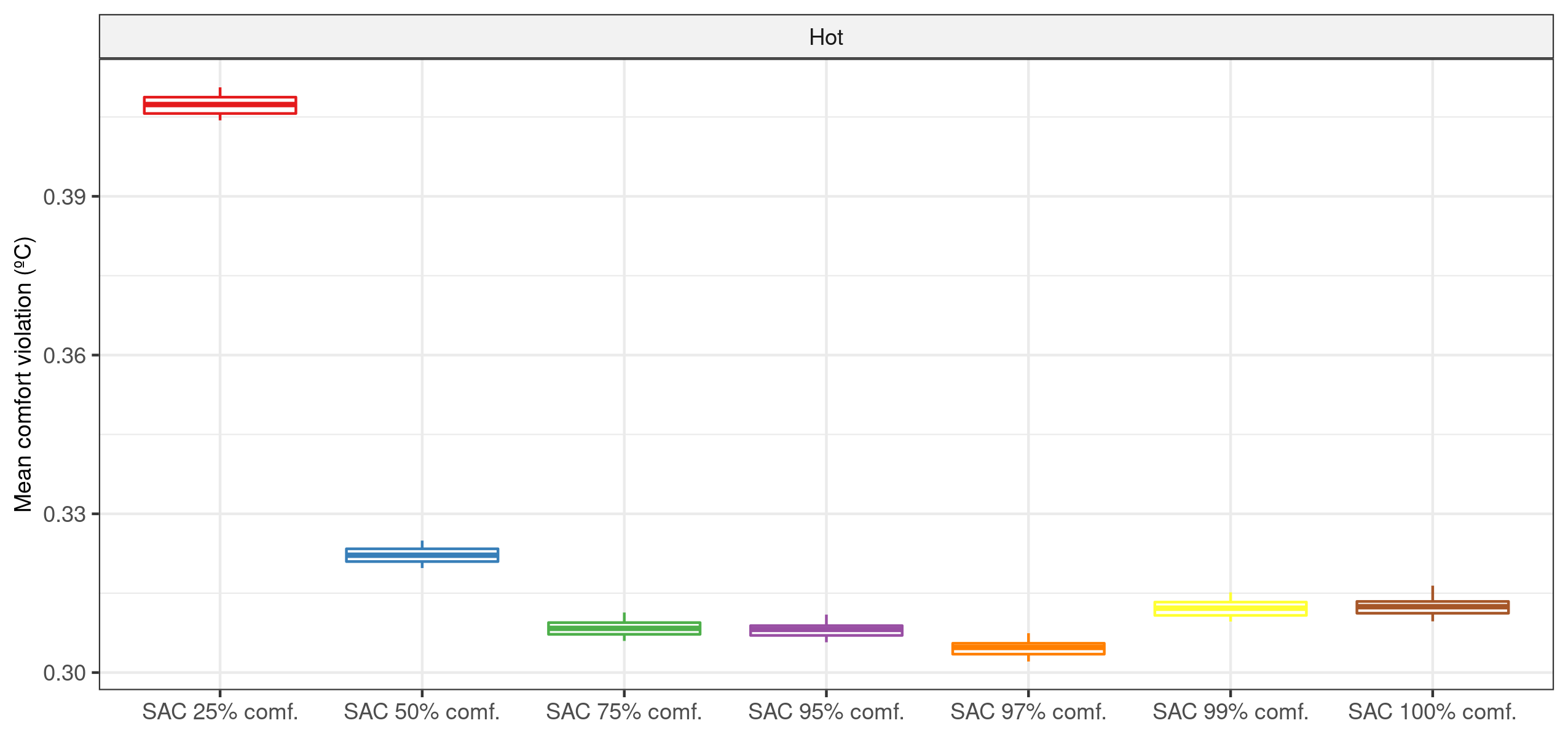}
    \caption{Mean episode comfort violations for SAC with additional comfort weights in \fivezone{}}
    \label{fig:mean_comf_tradeoff_SAC}
\end{figure}

Taking the SAC case as an example, if we look at Figure \ref{fig:mean_comf_tradeoff_SAC}, we observe that adding intermediate weights (95\%, 97\% and 99\%) to comfort between those previously considered (75\% and 100\%), leads to a range of similar results with centesimal variations. Therefore, the average comfort violations per episode reveal that there is some convergence in agents' performance, and that, despite increasing the importance of comfort in the reward function, their capacity is limited and it is not possible for them to improve comfort assurance. This is also the case for TD3 in \datacenter{} with cool weather, for which similar experiments were conducted.

Finally, we conclude that these two exceptions are indeed attainable cases which do not denote a noticeable loss of performance, but rather an oscillation around a limit that the agent is physically unable to overcome.

\subsection{Summary and implications}

We conclude with an overview of the results obtained and their potential implications for researchers and practitioners.

The results presented in sections \ref{sec:5.2} and \ref{sec:6.1} show how DRL-based agents match the performance of RBCs in simple scenarios, such as \fivezone{} ---where there is little room for improvement--- and how they perform better in environments of increasing complexity, such as \datacenter{}. We therefore recognise that the margin for improvement of a DRL over a reactive controller should be examined in the light of the complexity of the building being controlled. This complexity can be given by the system to be controlled, including its number of zones, as well as the observed and control variables. Other aspects may include climate variability or external factors such as occupancy or scheduling, which greatly affect the scalability and complexity of RBCs.

\vspace{0.5cm}
\begin{mdframed} \begin{quote}
DRL agents were able to level reactive controllers in simple environments and surpass them in complex environments. Therefore, we consider that the complexity of the building being controlled is a major factor to be taken into account in the selection of controllers.
\end{quote} \end{mdframed}
\vspace{0.5cm}

The robustness of agents deployed in environments with a climate different from that in which they were trained is another relevant issue (see sections \ref{sec:5.3} and \ref{sec:6.2}). Although there were yield losses, we believe that it will depend on the case to consider whether or not these losses are negligible.

On the other hand, we observed in sections \ref{sec:5.4} and \ref{sec:6.3} that a sequential learning approach did not show improvements, but biases towards the last environment used to train the agent. While it may be due to overfitting, the usefulness of this type of training needs to be further studied.

\vspace{0.5cm}
\begin{mdframed} \begin{quote}
Deploying a controller in an environment other than the one used in its training resulted in performance losses that might be acceptable depending on the case study. These losses were higher, and possibly impractical, in agents trained progressively through different climates, showing biases towards the last training environment employed.
\end{quote} \end{mdframed}
\vspace{0.5cm}

One of the main advantages that DRL controllers offer compared to reactive control techniques is their flexibility in customizing the comfort-consumption trade-off by simply altering their reward function. In sections \ref{sec:5.5} and \ref{sec:6.4}, we observed where are the limits of the comfort/consumption strategy that an agent can learn.

Moreover, while only these two factors have been considered in this paper, there are interesting studies, such as \cite{brandi2022}, where the estimation of power consumption and prices are taken into account, considering a DRL agent adapted to sporadic peak consumption and stationary events. This reveals the customization capabilities of DRL agents in building environments without a complex model of each controller or an extensive rule base.

\vspace{0.5cm}
\begin{mdframed} \begin{quote}
DRL agents are flexible to balance comfort and consumption preferences, and can incorporate other criteria by just adding extra terms in their reward.
\end{quote} \end{mdframed}
\vspace{0.5cm}

Facing the use of DRL controllers in real environments, we must consider that the variables available in the simulation are also available in the deployment environment. Most of the variables used in this study could be easily monitored in a real environment, or replaced by proxies if not available (e.g. comfort metrics).

Purely online deployment and training of these DRL agents in real environments is not recommended since it requires long periods of time to achieve a minimally efficient behavior policy. On the contrary, the approach usually followed is to train the agents in simulated environments, avoiding cold starts, and then deploy these agents in real environments and continue to receive feedback from the information collected online.

\vspace{0.5cm}
\begin{mdframed} \begin{quote}
In order to enable the deployment of a DRL agent trained in a simulated environment, it is necessary for the agent to have access, either directly or indirectly, to similar variables to those used in its training. The DRL setup facilitates continuous learning and adaptation after deployment.
\end{quote} \end{mdframed}
\vspace{0.5cm}


\section{Conclusions and future work}
\label{sec:7}

In this paper, we have addressed the performance comparison of multiple DRL algorithms in building HVAC control. Using the \sinergym{} framework, these algorithms were tested in various buildings and climates, considering both the search for comfort-consumption balance and scenarios in which one objective prevailed over the other. The robustness of the agents in different climates was also evaluated, as well as the potential benefits of progressive training to improve agents' performance.

In the search for proper comfort-consumption balance, SAC and TD3 were able to perform similar control to reactive controllers in most scenarios, even improving power savings at the cost of low comfort penalties. Since the agents were trained with completely raw observations ---neither exhaustive variable selection nor preprocessing beyond normalization--- and without extensive hyperparametrization, an immediate future work may involve the detailed study of which variables should compose agents' observations, as well as a deeper analysis to define which hyperparameters best fit each building-climate-algorithm combination. In any case, a relevant highlight derived from these results is the fact that positive results can be obtained for arbitrary environments without the need to significantly fine-tune observations and hyperparameters.

Regarding the robustness and generalisability of DRL agents, it was found that modifying the evaluation climate of an agent with respect to the one used in its training negatively influences its performance. Interestingly enough, these are minor losses that may be generally acceptable, avoiding the need to use a wide variety of climates in the training of the agents. Thus, a proposal for future work will involve training by sampling a pool of observations from a wide variety of climates, thus increasing the generalization capabilities of the agents and reducing training time and efforts. 

Moreover, the use of sequential learning for training was not successful in most cases, mainly due to the bias of the agents to specialize in the control of the most recently learned environments. This phenomenon of catastrophic forgetting motivates future research from different perspectives, such as exploring the influence of the network architecture of DRL agents on the occurrence of this phenomenon, or studying possible improvements along this progressive learning process (i.e. experience replay and regularization).

Finally, the study of the comfort-consumption trade-off, and its translation into the agents' reward function, offered some significant highlights. On the one hand, it was observed that there are cases where improvements in comfort have hardly any penalties in power consumption, which leads us to suggest that it cannot be assumed that increasing the importance of comfort in the reward function will always lead to a detriment in power demand. This is where reward engineering comes into play, so future research should be conducted to compare different rewards and their influence on the comfort-consumption trade-off. On the other hand, it was observed that there are comfort weights beyond which agents are physically incapable of improving temperature assurance in the desired ranges. Generally, as the comfort weight tended towards 100\% in the reward function, we began to find some convergence in the comfort penalties without considerable detriments or improvements.

As a conclusion of this work, and in view of the state of the art, we encourage future contributions in the area to be tested under multiple environments and configurations. Frameworks like \sinergym{} can be very useful in this regard. 

Addressing the problem of building energy optimization from the perspective of multi-agent systems is another interesting approach for which standardized comparisons are also necessary. These solutions offer several advantages in large buildings where coordination between multiple independent HVAC controllers is required \cite{nagarathinam2020marco,yu2020multi}.

From a wider perspective, it would also be interesting to extend these studies with configurations where the use of green energies is considered as a variable to be optimized, given the rise of smart grids and renewable energies. For this purpose, the continued use of common evaluation frameworks and methodologies will enable the comparison of results, the standardization of the field, and the joint progress towards the improvement of building energy optimization.



    
    
    
    

\backmatter

\section*{Declarations}

\bmhead{Funding}

Acknowledgements to grants SINERGY (PID2021.125537NA.I00) funded by MICIU/AEI/10.13039/501100011033 and ``ERDF -- A way of making Europe'', IA4TES (MIA.2021.M04.0008) funded by the Spanish Ministry of Economic Affairs and Digital Transformation (NextGenerationEU), and D3S (P21.00247), ``SE2021 UGR IFMIF-DONES'' funded by ERDF/Junta de Andalucía.

\bmhead{Conflict of interest}

The authors declare that they have no known competing financial or non-financial interests that could have appeared to influence the work reported in this paper.

\bmhead{Code and data availability}

The code and datasets used in this work are available in the following repository: \url{https://github.com/ugr-sail/paper-drl_building}.

\bmhead{Authors' contributions}

\textbf{Conceptualization}, \textbf{methodology}, \textbf{formal analysis} and \textbf{investigation}: all authors; \textbf{Data curation}: A. Manjavacas; \textbf{Software}: A. Campoy-Nieves, J. Jiménez-Raboso, A. Manjavacas; \textbf{Resources}: M. Molina-Solana, J. Gómez-Romero.  \textbf{Writing - original draft preparation}: A. Manjavacas; \textbf{Writing - review and editing}: Miguel Molina-Solana, Juan Gómez-Romero; \textbf{Visualization}: A. Manjavacas; \textbf{Funding acquisition}, \textbf{project administration} and \textbf{supervision}: M. Molina-Solana, J. Gómez-Romero.


\FloatBarrier

\begin{appendices}

\section{Hyperparameters}
\label{sec:appendixA}

Tables \ref{tab:hyp-ppo}, \ref{tab:hyp-sac} and \ref{tab:hyp-td3} show the hyperparameter combinations tested for each DRL algorithm. While for PPO and TD3 those marked in bold are the ones that provided the best performance, in the case of SAC all the combinations presented similar performance, so the default values of Stable Baselines3 were selected. We are aware of the complexity underlying the search and selection of hyperparameters in DRL, as there are many possible heuristics in the search for optimal values. However, this task is beyond the scope of this paper, and we will intend to address it in future research.

All of the combinations were run under the same random seed for a total of 20 training episodes, followed by 20 evaluation episodes, whose results were used to compare.

Regarding the architecture of the agents' networks, the default architectures provided by StableBaselines3 were used. These networks consist of 2 fully connected layers with:

\begin{itemize}
    \item 64 units per layer for PPO.
    \item 256 units for SAC.
    \item 400 (first layer) and 300 (second layer) units for TD3.
\end{itemize}

\begin{table}[htb]
    \caption{Hyperparameters tested for PPO}
    \label{tab:hyp-ppo}
    \footnotesize
    \centering
    \begin{tabular}{lllllll}
        \hline
          & \textbf{learning\_rate} & \textbf{n\_steps} & \textbf{batch\_size} & \textbf{n\_epochs} & \textbf{gamma} & \textbf{gae\_lambda}  \\ 
        \hline
        1 & 0.0003                  & 2048              & 64                   & 10                 & 0.99           & 0.95                  \\
        2 & 0.0003                  & 2048              & 128                  & 10                 & 0.99           & 0.95                  \\
        3 & 0.0003                  & 4096              & 128                  & 10                 & 0.99           & 0.95                  \\
        4 & 0.001                   & 2048              & 64                   & 10                 & 0.99           & 0.95                  \\
        5 & 0.01                    & 2048              & 64                   & 10                 & 0.99           & 0.95                  \\
        6 & \textbf{0.001}          & \textbf{4096}     & \textbf{128}         & \textbf{15}        & \textbf{0.9}   & \textbf{0.9}          \\
        7 & 0.0003                  & 4096              & 64                   & 15                 & 0.99           & 0.95                  \\
        \hline
    \end{tabular}
\end{table}

\begin{table}[htb]
    \caption{Hyperparameters tested for SAC}
    \label{tab:hyp-sac}
    \footnotesize
    \centering
    \begin{tabular}{llllll}
        \hline
          & \textbf{learning\_rate} & \textbf{buffer\_size} & \textbf{batch\_size} & \textbf{tau} & \textbf{gamma}  \\ 
        \hline
        1 & \textbf{0.0003}                  & \textbf{1000000}               & \textbf{256}                  & \textbf{0.005}        & \textbf{0.99}            \\
        2 & 0.003                   & 1000000               & 256                  & 0.005        & 0.99            \\
        3 & 0.0003                  & 10000                 & 256                  & 0.005        & 0.99            \\
        4 & 0.0003                  & 1000000               & 128                  & 0.005        & 0.99            \\
        5 & 0.0003                  & 1000000               & 256                  & 0.005        & 0.9             \\
        6 & 0.0003                  & 1000000               & 256                  & 0.05         & 0.99            \\
        7 & 0.0001                  & 100000                & 256                  & 0.003        & 0.9             \\
        \hline
    \end{tabular}
\end{table}

\begin{table}
    \caption{Hyperparameters tested for TD3. Combination 11 performed best for all environments, with the exception of \texttt{datacenter-mixed}, where combination 1 provided better results}
    \label{tab:hyp-td3}
    \footnotesize
    \centering
    \begin{tabular}{llllll}
        \hline 
                    & \textbf{learning\_rate} & \textbf{buffer\_size} & \textbf{batch\_size} & \textbf{tau}   & \textbf{gamma}  \\ 
        \hline
        1           & \textbf{0.003}                   & \textbf{1000000}               & \textbf{100}                  & \textbf{0.005}          & \textbf{0.9}             \\
        2           & 0.003                   & 1000000               & 128                  & 0.005          & 0.9             \\
        3           & 0.003                   & 100000000             & 100                  & 0.005          & 0.99            \\
        4           & 0.001                   & 1000000               & 100                  & 0.004          & 0.9             \\
        5           & 0.003                   & 10000000              & 100                  & 0.005          & 0.8             \\
        6           & 0.001                   & 100000                & 128                  & 0.005          & 0.8             \\
        7           & 0.003                   & 100000                & 100                  & 0.005          & 0.8             \\
        8           & 0.003                   & 1000000               & 256                  & 0.005          & 0.8             \\
        9           & 0.003                   & 1000000               & 512                  & 0.005          & 0.99            \\
        10          & 0.003                   & 1000000               & 128                  & 0.005          & 0.7             \\
        11\textbf{} & \textbf{0.003}          & \textbf{1000000}      & \textbf{64}          & \textbf{0.005} & \textbf{0.9}    \\
        \hline
    \end{tabular}
\end{table}

\FloatBarrier

\section{Evaluation results}
\label{sec:appendixB}

\edited{Tables \ref{tab:eval-5zone-values} and \ref{tab:eval-datacenter-values} contain the numerical results in terms of reward, power consumption, and comfort for the evaluated agents. The row names refer to:}

\begin{itemize}
    \item \textbf{Rew.}: mean episode reward.
    \item \textbf{Pow.}: mean episode power consumption (W).
    \item \textbf{Comf.}: mean episode comfort violation time (\%).
\end{itemize}

\begin{table}
    \caption{\edited{Mean values and standard deviations for evaluation metrics in \fivezone{}}}
    \label{tab:eval-5zone-values}
    \resizebox{\textwidth}{!}{%
    \centering
    \begin{tabular}{llll} 
        \hline
                        & \textbf{Cool}                                                                                                                                                                                                                  & \textbf{Mixed}                                                                                                                                                                                                                 & \textbf{Hot}                                                                                                                                                                                                                    \\ 
        \hline
        \textbf{Rew.}  & \begin{tabular}[c]{@{}l@{}}\textbf{PPO} (mean = -0.250, sd = 0.001)\\ \textbf{SAC} (mean = -0.226, sd = 0.001)\\ \textbf{TD3} (mean = -0.258, sd = 0.001)\\ \textbf{RBC} (mean = -0.198, sd = 0.003)\\ \textbf{RAND} (mean = -0.601, sd = 0.002)\end{tabular}                & \begin{tabular}[c]{@{}l@{}}\textbf{PPO} (mean = -0.367, sd = 0.002)\\ \textbf{SAC} (mean = -0.344, sd = 0.002)\\ \textbf{TD3} (mean = -0.368, sd = 0.002)\\ \textbf{RBC} (mean = -0.284, sd = 0.001)\\ \textbf{RAND} (mean = -0.715, sd = 0.002)\end{tabular}                & \begin{tabular}[c]{@{}l@{}}\textbf{PPO} (mean = -0.470, sd = 0.003)\\ \textbf{SAC} (mean = -0.454, sd = 0.001)\\ \textbf{TD3} (mean = -0.477, sd = 0.001)\\ \textbf{RBC} (mean = -0.416, sd = 0.001)\\ \textbf{RAND} (mean = -0.735, sd = 0.002)\end{tabular}                 \\ 
        \hline
        \textbf{Pow.}   & \begin{tabular}[c]{@{}l@{}}\textbf{PPO} (mean = 3315.004, sd = 22.956)\\ \textbf{SAC} (mean = 3161.168, sd = 22.741)\\ \textbf{TD3} (mean = 3439.188, sd = 23.281)\\ \textbf{RBC} (mean = 3317.391, sd = 24.349)\\ \textbf{RAND} (mean = 2476.047, sd = 17.468)\end{tabular} & \begin{tabular}[c]{@{}l@{}}\textbf{PPO} (mean = 5506.539, sd = 38.463)\\ \textbf{SAC} (mean = 5248.838, sd = 35.758)\\ \textbf{TD3} (mean = 5681.867, sd = 38.989)\\ \textbf{RBC} (mean = 5642.405, sd = 32.619)\\ \textbf{RAND} (mean = 4207.539, sd = 27.431)\end{tabular} & \begin{tabular}[c]{@{}l@{}}\textbf{PPO} (mean = 5949.604, sd = 15.845)\\ \textbf{SAC} (mean = 5860.091, sd = 16.135)\\ \textbf{TD3} (mean = 6260.862, sd = 16.208)\\ \textbf{RBC} (mean = 5893.126, sd = 13.720)\\ \textbf{RAND} (mean = 4900.652, sd = 13.134)\end{tabular}  \\ 
        \hline
        \textbf{Comf.} & \begin{tabular}[c]{@{}l@{}}\textbf{PPO} (mean = 32.997, sd = 0.137)\\ \textbf{SAC} (mean = 28.254, sd = 0.168)\\ \textbf{TD3} (mean = 33.972, sd = 0.066)\\ \textbf{RBC} (mean = 26.416, sd = 0.379)\\ \textbf{RAND} (mean = 63.044, sd = 0.295)\end{tabular}                & \begin{tabular}[c]{@{}l@{}}\textbf{PPO} (mean = 33.661, sd = 0.109)\\ \textbf{SAC} (mean = 33.101, sd = 0.138)\\ \textbf{TD3} (mean = 33.429, sd = 0.005)\\ \textbf{RBC} (mean = 27.791, sd = 0.369)\\ \textbf{RAND} (mean = 61.312, sd = 0.256)\end{tabular}                & \begin{tabular}[c]{@{}l@{}}\textbf{PPO} (mean = 33.805, sd = 0.113)\\ \textbf{SAC} (mean = 32.853, sd = 0.144)\\ \textbf{TD3} (mean = 34.716, sd = 0.085)\\ \textbf{RBC} (mean = 30.441, sd = 0.256)\\ \textbf{RAND} (mean = 58.166, sd = 0.195)\end{tabular}                 \\
        \hline
        \end{tabular}
    }
\end{table}
    
\begin{table}
    \caption{\edited{Mean values and standard deviations for evaluation metrics in \datacenter{}}}
    \label{tab:eval-datacenter-values}
    \resizebox{\textwidth}{!}{%
    \centering
    \begin{tabular}{llll} 
        \hline
                       & \textbf{Cool}                                                                                                                                                                                                                   & \textbf{Mixed}                                                                                                                                                                                                                       & \textbf{Hot}                                                                                                                                                                                                                          \\ 
        \hline
        \textbf{Rew.}  & \begin{tabular}[c]{@{}l@{}}\textbf{PPO} (mean = -0.406, sd = 0.004)\\ \textbf{SAC} (mean = -0.352, sd = 0.004)\\ \textbf{TD3} (mean = -0.326, sd = 0.004)\\ \textbf{RBC} (mean = -0.331, sd = 0.002)\\ \textbf{\textbf{RAND}} (mean = -0.879, sd = 0.006)\end{tabular}                & \begin{tabular}[c]{@{}l@{}}\textbf{PPO} (mean = -0.948, sd = 0.004)\\ \textbf{SAC} (mean = -0.934, sd = 0.003)\\ \textbf{TD3} (mean = -0.856, sd = 0.004)\\ \textbf{RBC} (mean = -0.877, sd = 0.005)\\ \textbf{\textbf{RAND}} (mean = -1.424, sd = 0.005)\end{tabular}                     & \begin{tabular}[c]{@{}l@{}}\textbf{PPO} (mean = -1.363, sd = 0.003)\\ \textbf{SAC} (mean = -1.299, sd = 0.003)\\ \textbf{TD3} (mean = -1.216, sd = 0.003)\\ \textbf{RBC} (mean = -1.282, sd = 0.004)\\ \textbf{\textbf{RAND}} (mean = -1.830, sd = 0.006)\end{tabular}                      \\ 
        \hline
        \textbf{Pow.}  & \begin{tabular}[c]{@{}l@{}}\textbf{PPO} (mean = 8036.334, sd = 78.246)\\ \textbf{SAC} (mean = 6933.573, sd = 74.947)\\ \textbf{TD3} (mean = 6254.018, sd = 72.473)\\ \textbf{RBC} (mean = 6620.088, sd = 38.331)\\ \textbf{\textbf{RAND}} (mean = 7271.774, sd = 88.468)\end{tabular} & \begin{tabular}[c]{@{}l@{}}\textbf{PPO} (mean = 18893.892, sd = 74.454)\\ \textbf{SAC} (mean = 18382.084, sd = 62.678)\\ \textbf{TD3} (mean = 16854.495, sd = 79.335)\\ \textbf{RBC} (mean = 17547.968, sd = 97.998)\\ \textbf{\textbf{RAND}} (mean = 18212.941, sd = 91.845)\end{tabular} & \begin{tabular}[c]{@{}l@{}}\textbf{PPO} (mean = 26861.237, sd = 63.086)\\ \textbf{SAC} (mean = 25453.090, sd = 60.553)\\ \textbf{TD3} (mean = 24022.427, sd = 65.902)\\ \textbf{RBC} (mean = 25638.081, sd = 73.229)\\ \textbf{\textbf{RAND}} (mean = 25992.284, sd = 74.121)\end{tabular}  \\ 
        \hline
        \textbf{Comf.} & \begin{tabular}[c]{@{}l@{}}\textbf{PPO} (mean = 2.009, sd = 0.063)\\ \textbf{SAC} (mean = 4.823, sd = 0.157)\\ \textbf{TD3} (mean = 29.883, sd = 0.267)\\ \textbf{RBC} (mean = 0.003, sd = 0)\\ \textbf{\textbf{RAND}} (mean = 64.582, sd = 0.213)\end{tabular}                       & \begin{tabular}[c]{@{}l@{}}\textbf{PPO} (mean = 1.819, sd = 0.100)\\ \textbf{SAC} (mean = 7.303, sd = 0.143)\\ \textbf{TD3} (mean = 17.344, sd = 0.235)\\ \textbf{RBC} (mean = 0.003, sd = 0)\\ \textbf{\textbf{RAND}} (mean = 64.288, sd = 0.350)\end{tabular}                            & \begin{tabular}[c]{@{}l@{}}\textbf{PPO} (mean = 6.697, sd = 0.142)\\ \textbf{SAC} (mean = 13.151, sd = 0.225)\\ \textbf{TD3} (mean = 14.910, sd = 0.248)\\ \textbf{RBC} (mean = 0.003, sd = 0)\\ \textbf{\textbf{RAND}} (mean = 65.026, sd = 0.234)\end{tabular}                            \\
        \hline
    \end{tabular}
    }
\end{table}

\end{appendices}


\FloatBarrier

\bibliography{sn-bibliography}

\end{document}